%% file: 2018-ModeInference_ARXIV_FINAL.tex
\documentclass{article}

\usepackage[preprint]{nips_2018}
\usepackage{natbib}
\setcitestyle{round}
\usepackage[utf8]{inputenc} 
\usepackage[T1]{fontenc}    
\usepackage{hyperref}       
\usepackage{url}            
\usepackage{booktabs}       
\usepackage{amsfonts}       
\usepackage{nicefrac}       
\usepackage{microtype}      

\usepackage{amsmath}
\usepackage{bm}
\usepackage{amssymb}
\usepackage{upgreek}
\usepackage{mathtools}
\usepackage{booktabs}
\usepackage{tabu}
\usepackage{accents}

\usepackage{color}

\input{mydefs}

\renewcommand{\vec}[1]{\mathbf{#1}}
\def\netloss{\mathcal{L}}
\newcommand{\emptycell}{-}

\def\METHOD*{REPRISE}

\title{Learning, Planning, and Control in a Monolithic Neural Event Inference Architecture}

\author{Martin~V. Butz\textsuperscript{1} \And David Bilkey\textsuperscript{2a} \And Dania Humaidan\textsuperscript{1} \And Alistair Knott\textsuperscript{2b} \And Sebastian Otte\textsuperscript{1}\\\\
	\textsuperscript{1}Cognitive Modeling Group\\
	Computer Science Department\\
	University of Tübingen\\
	Sand 14, 72076 Tübingen, Germany\\\\
	\textsuperscript{2}Department of ${{}^{a}}$Psychology / ${{}^{b}}$Computer Science\\
	University of Otago\\
	P.O. Box 56, Dunedin, New Zealand\\
}

\begin{document}
	
	\maketitle

\begin{abstract}
We introduce \METHOD*,
a REtrospective and PRospective Inference SchEme,
which learns temporal event-predictive models of dynamical systems. 
\METHOD* infers the unobservable contextual event state and accompanying temporal predictive models that best explain the recently encountered sensorimotor experiences {\em retrospectively}. 
Meanwhile, it optimizes upcoming motor activities {\em prospectively} in a goal-directed manner.
Here, \METHOD* is implemented by a recurrent neural network (RNN), which learns temporal forward models of the sensorimotor contingencies generated by different simulated dynamic vehicles.
The RNN is augmented with contextual neurons, which enable the encoding of distinct, but related, sensorimotor dynamics as compact event codes. 
We show that \METHOD* concurrently learns to separate and approximate the encountered sensorimotor dynamics:
it analyzes sensorimotor error signals adapting both internal contextual neural activities and connection weight values.
Moreover, we show that \METHOD* can exploit the learned model to induce goal-directed, model-predictive control, that is, approximate active inference:  
Given a goal state, the system imagines a motor command sequence optimizing it with the prospective objective to minimize the distance to the goal.
The RNN activities thus continuously imagine the upcoming future and reflect on the recent past, optimizing the predictive model, the hidden neural state activities, and the upcoming motor activities.
As a result, event-predictive neural encodings develop, which allow the invocation of highly effective and adaptive goal-directed sensorimotor control.
\end{abstract}


\section{Introduction}

The predictive brain perspective and active inference principles have strongly influenced cognitive science over the last years \citep{Bar:2009,Butz:2017,Clark:2016,Friston:2009,Hohwy:2013}. 
Although predictive encodings have shown to yield promising results in artificial neural networks focusing on vision \citep{Rao:1999}, it remains highly challenging to realize these principles in scalable, \emph{temporal} dynamic artificial neural network models, and particularly models that enable flexible, goal-directed planning (but see \citealp{Najnin:2017} for a recent, promising approach in phonological speech production). 
Moreover, it remains unclear how abstracted, hierarchical structures may be developed effectively \citep{Botvinick:2014,McClelland:2010} -- structures that are believed to be essential for enabling the generation of flexible, adaptive goal-directed behavior by means of hierarchical, model-based planning and reinforcement learning \citep{Botvinick:2009}.

Despite the recent remarkable successes in playing somewhat challenging computer games and the board game GO \citep{Mnih:2015,Silver:2016}, neural networks generally still seem to lack a deeper understanding of the underlying problem domain. 
As a result, the developed systems are rather inflexible, for example, when multiple, different tasks need to be solved by the same architecture or when the reward function changes. 
There are certainly strategies that can help --- such as more effective episodic replay or task-specific weight and neural manipulations \citep{Kirkpatrick:2016}. 
Nonetheless, deep and recurrent neural networks do not yet learn and think with the flexibility of humans \citep{Lake:2017}.

We introduce a novel retrospective and prospective temporal inference scheme (\METHOD*), which we implement in recurrent artificial neural networks (RNNs).
\METHOD* combines weight adaptation (i.e. model inference) with neural activity adaptation (i.e. contextual hidden state inference) and model-predictive, anticipatory control and behavior (i.e. active inference). 
As a result, \METHOD* does not optimize a static reward function, but it can flexibly plan context- and task-dependently. 
Moreover, \METHOD* offers a first step towards the development of hierarchical hidden, generative structures, which can be closely related to the concept of event cognition.

The event cognition principle comes from cognitive psychology.
It was shown that humans have a strong tendency to segment a continuous sensorimotor stream into meaningful events and event-transitions, leading to the proposal of an event segmentation theory (EST)
\citep{Radvansky:2014,Zacks:2001,Zacks:2007}.
Concurrently, the theory of event coding (TEC) has proposed integrative action-effect codes, referring to them as event codes \citep{Hommel:2001}.
Similarly, forward-inverse control schemes have been put forward as models of human behavior \citep{Wolpert:1998,Wolpert:2016}, where the involved forward-inverse models essentially encode interaction events.
Even the memorization of experienced episodes appears event-segmented and event-focused \citep{Richmond:2017}. 
Moreover, memorized events can be used not only for processing current sensorimotor information, but also for reflecting on the past, for imagining potential futures, or even for reasoning about fully hypothetical events \citep{Bar:2009,Buckner:2007,Schacter:2012}.
Combined with the predictive coding perspective on cognition, our mind appears to have the tendency to cluster sensorimotor contingencies into predictive events \citep{Butz:2016}.

\METHOD* offers an RNN-based neural implementation that shows the emergent tendency to cluster different types of predictively encoded sensorimotor dynamics into compact event codes without the provision of event-type or event-boundary information.
Related work with RNNs has implemented hierarchical RNN architectures that develop symbol-like encodings in the deeper RNN layer via gradient descent \citep{Tani:2003}. 
Later, the process was termed an error regression scheme and was closely related to the free energy principle \citep{Murata:2017}. 
Related work has also been put forward when setting internal hidden states --- often referred to as \emph{parametric bias} neurons --- to induce particular behavioral primitives and sequences thereof by suitably trained hierarchical RNN architectures \citep{Arie:2009,Tani:1996a,Sugita:2011a}.
Interestingly, the activities of the parametric bias neurons were shown to exhibit compositional, pre-linguistic structures \citep{Sugita:2011,Sugita:2011a}, which were linked with a language production system elsewhere \citep{Sugita:2005}.

\METHOD* combines goal-directed, active-inference-based control with the learning of stable hidden contextual states---without the help of episodic training, the provision of interaction types, or boundary signals between interactions. 
We relate the inference of these hidden states with event-predictive encodings, which tend to contrast distinct sensorimotor dynamics.
Somewhat similar stable states have been inferred recently in a spiking neural network architecture, where free energy minimization techniques were applied to establish temporary bindings in a distributed network \citep{Pitti:2017}.
Moreover, stochastic search was used in these spiking networks to induce goal-directed behavior. 
In \METHOD*, context-adaptive goal-directed behavior is generated via gradient-based, prospective inference, which is closely related to \emph{active inference} with respect to the free energy formalism \citep{Friston:2009}.
As a result, \METHOD* adapts its internal state and its motor behavior in such a way that  approximately optimal goal-directed behavior is generated.
From a control perspective, the system can be said to approximate model-predictive control \citep{Camacho:1999a}, where the model is learned by and then encoded in an RNN.

As a first evaluation scenario, we train \METHOD* to control three different types of ``vehicles'' in a simple but dynamic 2D simulated environment. 
Crucially, we show that \METHOD* is able to distinguish the three types of vehicles and control them effectively in a goal-directed manner even when no information about the vehicle identity or even the fact that there are three different vehicles is provided. 
We believe that this method may be very well-suited to learn event-oriented abstractions and event hierarchies, but future work is necessary to scale the system and apply it to more challenging scenarios.

\section{System Architecture and Inference Mechanisms}

We now detail the mechanisms implemented by our retrospective and prospective temporal inference scheme (\METHOD*), which is  implemented in an RNN. 
\METHOD* infers the unobservable current event context (here the controlled vehicle), which best explains the recent sensorimotor experiences, retrospectively.
Meanwhile, it infers motor control commands prospectively in a goal-directed manner. 
We thereby build on our previous work, which had accomplished prospective, active motor control inference \citep{Otte:2017,Otte:2017a} but not retrospective inference.
Our results suggest that \METHOD* can learn and apply both, effective goal-directed control and event-oriented, system state inference.

In order to introduce \METHOD*, we distinguish between the actual (not directly observable) dynamical system $\phi$ and the model $\Phi$ of this system, which is encoded by an RNN.
Focusing on a discrete-time dynamical system, at a certain point in time $t$, the (not directly observable) current state of the dynamical system $\phi$ may be denoted by $\boldsymbol{\vartheta}^t$, such that the progression through time is determined by   
\begin{equation}
\boldsymbol{\vartheta}^t \xmapsto{\displaystyle~\phi~} \boldsymbol{\vartheta}^{t+1}.
\end{equation}

\subsection{Temporal Forward Model}
\begin{figure*}[t]
	\centering
	\includegraphics[width=0.6 \textwidth]{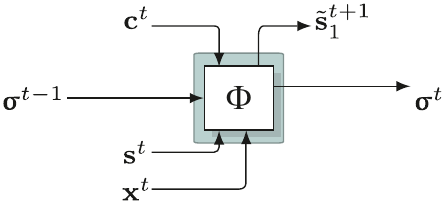}
	\caption{
		Illustration of the recurrent, temporal predictive sensorimotor forward model $\Phi$'s input to output processing activities, including neural contextual signals $\vec{c}^t$.}    
	\label{figure:wiring}
\end{figure*}

The model $\Phi$ is trained to approximate these dynamics, inferring its parameters from sensorimotor experiences during learning.
However, seeing that we are dealing with a dynamic, \emph{partially observable Markov decision process} (POMDP) \citep{Sutton:1998}, the true system state $\boldsymbol{\vartheta}^t$ is typically not directly deducible from current observables $\vec{s}^{t} \in \mathbb{R}^{n}$.
Thus, the dynamical system's internal state $\boldsymbol{\sigma}^{t}$ must be inferred in each iteration from the current observables $\vec{s}^{t}$, the current motor activities denoted by $\vec{x}^{t} \in \mathbb{R}^{k}$, and the previous system state estimate $\boldsymbol{\sigma}^{t-1}$.
With the help of the system's model $\Phi$, the next system state $\boldsymbol{\sigma}^{t+1}$ and the consequent sensory expectations $\tilde{\vec{s}}^{t+1}$ are determined by 
\begin{equation}\label{equation:forwardmodel}
(\vec{s}^{t}, \boldsymbol{\sigma}^{t-1}, \vec{x}^{t})
\xmapsto{\displaystyle~\Phi~} 
(\tilde{\vec{s}}^{t+1}, \boldsymbol{\sigma}^{t}),
\end{equation}
where the mapping $\Phi$ essentially models the temporal forward dynamics of the system.
Thus, the next system state and sensory expectations depend on the current sensor ($\vec{s}^{t}$) and motor control ($\vec{x}^{t}$) activities as well as, in principle, on the entire state history, which is encoded in the (hidden) state components ($\boldsymbol{\sigma}^{t-1}$) in compressed form.
Figure~\ref{figure:wiring} shows the input and output signals that are processed via the model $\Phi$.

While learning the model, that is, while pursuing model inference, the system attempts to minimize the squared loss between predicted and encountered sensory information over time, that is,
\begin{equation}
\netloss = \sum_{t=1}^{T} \sum_{i=1}^{n}\frac{1}{2}(\tilde{\vec{s}}_i^{t} - \vec{s}_i^{t})^2,
\label{eq:netloss} 
\end{equation} 
summing the accumulated losses over the gathered experiences $\{\vec{s}_1,\dots,\vec{s}_{T}\}$.

\subsection{Multiple Dynamical Systems}
In this paper we consider the challenge when not only a single dynamical system needs to be controlled, but an ensemble of multiple dynamical systems $\phi = \{\phi_{1},\ldots,\phi_{u}\}$, which cause different sensorimotor contingencies over time. 
These systems differ from each other concerning their behavior, but share the same input, state, and output dimensions. 
During model inference, the model $\Phi$ is trained to approximate all of these dynamical systems within one monolithic RNN architecture. 
As a result, the challenge is to approximate the particular dynamical system $\phi_{i}$ that is currently active, given observed state $\vec{s}^{t}$ and control commands $\vec{x}^{t}$, in order to be able to accurately predict future sensorimotor dynamics and thus to control the dynamical system itself in an anticipatory, goal-directed manner.

When evaluating \METHOD* we will first provide the identity of the currently active dynamical system $\phi_{i}$ in additional context input neurons $\vec{c} \in \mathbb{R}^{u}$, which is simply added as additional input and which is initially encoded as a one-hot vector ($i$-th component is set to $1$, rest to $0$).
This encoding is closely related to parametric bias neurons, which can be viewed as an indicator of the current event the system is situated in \citep{Sugita:2011a,Tani:2017}.
During goal-directed control, however, we will infer the values of this vector online. 
Later on, we will infer this vector also during training (never providing information about the current vehicle identity) and will show that the different vehicles tend to be encoded separately in the provided contextual neurons.  

%
%

\subsection{\METHOD*}\label{section:inverse}
Given an imagined action sequence, an initial state, and the identity of the current dynamical system, the RNN can predict a state progression that is expected when executing the action sequence by means of the learned temporal forward model $\Phi$.
To control the system effectively, however, the inverse mapping is required, that is, an action sequence needs to be inferred to approach a desired goal-state (or follow a sequence of goal-states) from an initial state. 
This becomes even more difficult when the identity of the current actual dynamical system $\phi_i$ is unknown and has to be inferred as well. 
In this section we introduce the \METHOD* algorithm --- a concurrent retrospective and prospective inference scheme, 
which solves the twofold system identification and goal-directed
control problem.

\begin{figure*}[t]
	\centering
	\includegraphics[width=\textwidth]{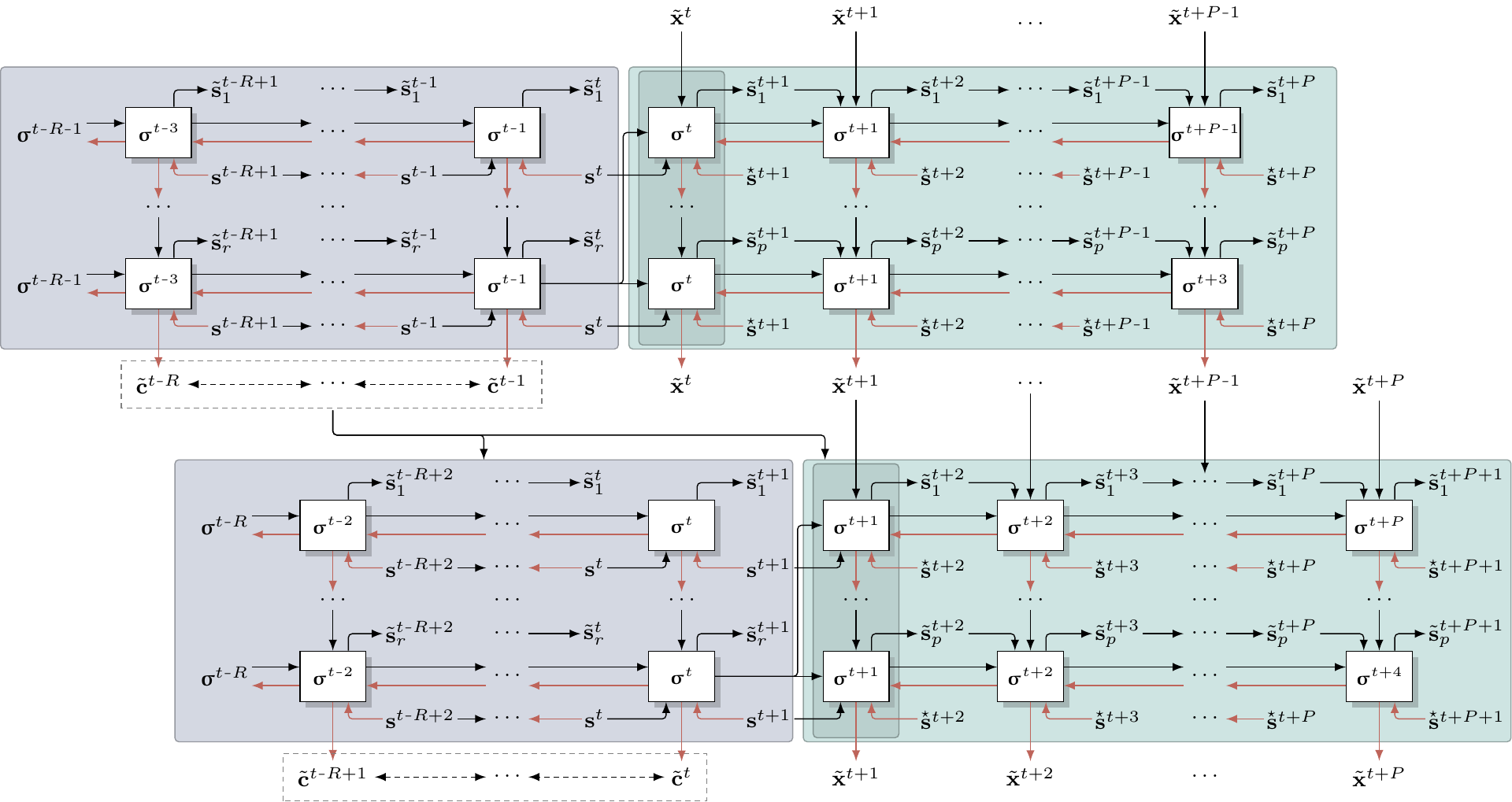}
	\caption{
		Illustration of \METHOD* for two consecutive time steps $t$ (top part) and $t+1$ (bottom part).
		Note that there is only one RNN, whose activities $\vec{\sigma}$ are buffered over time.
		The right (green shaded) boxes illustrate future imaginations actively inferring prospective motor activities, while the left boxes (gray shaded) show retrospections about the recent past for system state inference (including event state $\vec{c}^{t'}$ and hidden state $\smash{\boldsymbol{\sigma}^{t}}$).
		Black lines indicate context and information forward flow, while the red lines indicate gradient flow. 
		$\smash{\tilde{\vec{x}}^{t'}}$ and $\smash{\tilde{\vec{c}}^{t'}}$ refer to the action and context input vectors, respectively, for a particular time step $t'$. 
		$\smash{\tilde{\vec{s}}_{\tau}^{t'}}$ refers to a particular sensory prediction in the $\tau$-th optimization cycle, whereas $\smash{\accentset{\star}{\vec{s}}^{t'}}$ refers to a desired sensory goal state.}    
	\label{figure:inferenceapproach}
\end{figure*}

Figure~\ref{figure:inferenceapproach} shows the dynamic processes \METHOD* unfolds for two consecutive time steps. 
During each step, both a retrospective and a prospective inference phase is executed. 

In the retrospective phase, the gradient is propagated $R$ time steps into the past, to reflect on the states that were just experienced.
The gradient is fed by the discrepancy between previously predicted system states $\tilde{\vec{s}}^{t-i}$, with $i\in{0,\dots,R}$, and the actually observed system states $\vec{s}^{t-i}$, minimizing the quadratic loss over this time horizon (\ref{eq:netloss}). 
The discrepancy is then mapped onto the assumed context input $\vec{c}^{t-i}$ --- essentially a sub-vector of $\vec{s}^{t-i}$ --- indicating the dynamical system $\phi_i$ that is presumably currently active.
Because \METHOD* is designed to distinguish different events, $\vec{c}$ is set to a constant value over the time horizon $R$ --- essentially summing up the gradient signals received by $\vec{c}^{t-1}\dots \vec{c}^{t-R}$ when applying gradient descent. 
Additionally, the error gradient can be used to adapt the RNN's hidden state at time step $t-R-1$, that is, $\vec{\sigma}^{t-R-1}$, such that it better fits the changing context input. 
As a result, the RNN avoids disadvantageous or even undefined sensory input, motor command, hidden state combinations.
After neural activity adaptation via gradient descent with learning rates $\eta_c$ and $\eta_\sigma$, respectively (we apply Adam, \citealp{Kingma:2014}), the neural activities are propagated forward again to the present time step, with respect to the inferred hidden state and context input, and the already recorded motor commands and observed system states, yielding an updated $\vec{\sigma}^t$.
This retrospective neural activity inference cycle may be executed $r$ times.

Additionally, retrospective weight adaptation is applied during model learning, again focusing on minimizing the quadratic loss (\ref{eq:netloss}).
This essentially corresponds to standard back-propagation through time learning. 
Note that we usually use a deeper retrospective time horizon $R_w$ for model learning and a smaller learning rate $\eta_w$, when compared to $\eta_c$ and $\eta_\sigma$.
Otherwise the RNN would behave more like an adaptive filter as it would not learn the (versatile) model characteristics but rather over-fit the recent signal shape.

In the prospective phase, neural activities are projected $P$ time steps into the future, starting with the inferred current internal system state $\vec{\sigma}^t$ and hypothetically executing a sequence of motor commands $\smash{\tilde{\vec{x}}^{t+i}}$, which was inferred previously.
The discrepancies between the predicted future $\smash{\tilde{\vec{s}}^{t+i}}$ and desired goal state sequences $\smash{\accentset{\star}{\vec{s}}^{t+i}}$, with $i\in{1,\dots,P}$, are then propagated backwards through time from the imagined future back to the present time step, while the gradient is projected onto the individual 
anticipated motor activity sequence $\smash{\tilde{\vec{x}}^{t+i}}$, effectively optimizing it in the light of the current system state estimates and the desired goal state. 
This prospective inference cycle is executed $p$ times.

After the retro- and prospective inference phases, the inferred motor activity $\vec{x}^t$ is executed by the system $\phi$ and the forward RNN is updated via (\ref{equation:forwardmodel}), closing the processing loop. Then \METHOD* is repeated in the following time step ($t+1$).\footnote{\METHOD* Java code can be found in the following github repository:\\ \url{https://github.com/CognitiveModeling/2019-ModeInferencePaperCode}.}

\section{System Evaluations}
Our experiments are based on a two dimensional dynamical system simulation with $u=3$ types of ``vehicles'', constituting three dynamical systems: 
\begin{itemize}
\item $\phi_{1}$ is a multi-copter-like vehicle, which we call \emph{rocket},
\item $\phi_{2}$ is a static omnidirectional vehicle, which we call \emph{stepper}, and 
\item $\phi_{3}$ is a dynamical, omnidirectional gliding vehicle, which we call \emph{glider}.
\end{itemize}

The rocket is influenced by simulated gravity and undergoes inertia. 
It has two propulsion motors that are spread at a $45^{\circ}$\,angle from the vertical axis on both sides, inducing thrust forces in the respective direction.
The two other motor inputs are ignored in the case of the rocket.
The stepper has four thrust motors that are spread at $45^{\circ}$ and $135^{\circ}$\,angle from the vertical axis to both sides, inducing steps in the opposite direction.
Finally, the glider has the same four thrust motors as the stepper.
However, in contrast to the stepper, the glider undergoes inertia without any friction.  
Each motor unit can be throttled within the interval $[0, 1]$.
Upon invocation, each vehicle is positioned in a rectangular free space of size $3\times 2$ units.
It is surrounded by borders, which block the vehicle. 

All presented evaluations below are based on ten independently trained networks, averaging the achieved results.
Here, we evaluate \METHOD* when the context neuron activities are set during training to distinct one-hot vector values for the different vehicles.
In Section~\ref{sec:contextInference}, we then investigate learning and performance when context values are inferred during learning as well.

\subsection{Model Learning}
During training, stochastic back-propagation through time optimized the weights of the considered RNN architectures based on simulated sensorimotor experiences, learning in a self-supervised manner.
Experiences were generated by executing pseudo-random motor commands $\vec{x}\in[0, 1]^4$, where motor command generation was such that sufficient upwards thrust was generated and a reasonable exploration of the complete rectangular free space was loosely ensured.

\begin{figure}[t]
	\centering
	\includegraphics[width=\linewidth]{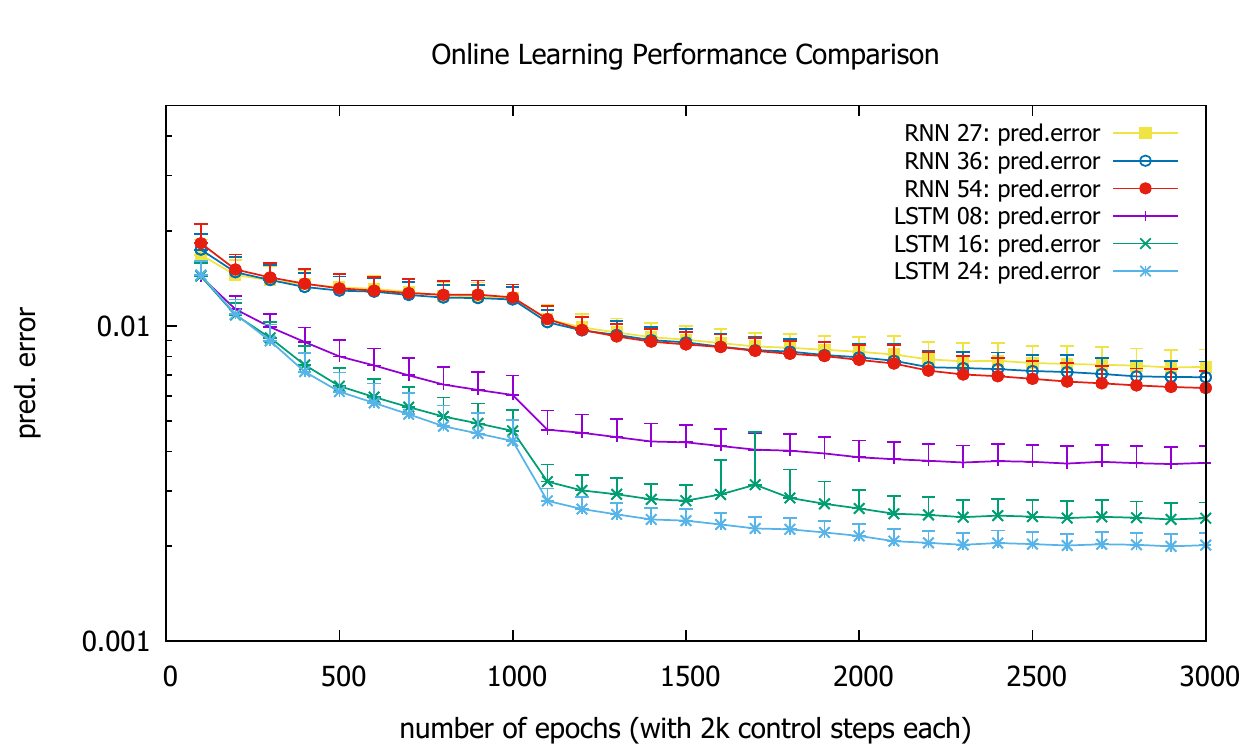}
	\caption{Learning progress comparing standard RNNs and LSTMs with different numbers of hidden units. Learning rate is reduced by a factor of $.1$ after 1k and 2k epochs.}    
	\label{figure:learning}
\end{figure}

At each time step 
the network is fed with the current position of the vehicle ($\vec{s}\in\{[-1.5,1.5],[0,2]\}$), the current four motor commands (activities of the four thrust motors as forces; for the rocket, the second two motor values have no effect; mass of vehicles is set to $.1$), and a three bit one-hot vector, which indicates the vehicle that is currently controlled, i.e., which $\phi_i$ applies. 
The network predicts the vehicle's resulting change in position.

We trained the considered RNNs in $3000$ independent epochs, consisting of $2000$ pseudo-random control steps each. 
We applied back-propagation through time every 50 iterations and Adam as the weight adaptation mechanism \citep{Kingma:2014}.
The learning rate was annealed, such that $\eta = .001$, $\eta = 10^{-4}$, $\eta = 10^{-5}$ during the first, second, and third 1000 epochs, respectively. (First and second moment smoothing factors were set to the standard values $\beta_{1}=0.9$, $\beta_{2}=0.999$.)
Each vehicle was simulated for $2000$ time steps (i.e. one epoch), after which the hidden state of the RNN was reset to zero and a new vehicle was initialized. 

Figure~\ref{figure:learning} contrasts the sensory prediction error development during learning for several RNN architectures, showing averaged mean errors and standard deviations across 20 independently weight-initialized (normally distributed values with standard deviation $0.1$) networks. 
Standard RNNs with one hidden layer of 27 (1026), 36 (1692), and 54 neurons (3510 weights) perform consistently worse than long short-term memory (LSTM) RNNs with forget gates and peephole connections \citep{Gers:2002}. 
While 16 hidden memory cells (1680 weights) clearly outperform 8 hidden memory cells (584 weights), the advantage of yet another 8 hidden cells, that is, 24 cells in total (3288 weights) is less pronounced. 

\begin{figure*}[hbt!]
	\centering
	\setlength{\fboxsep}{1pt}
	\vspace{-1.0cm}
	\fbox{\includegraphics[trim={9cm 5.6cm 10cm   6.5cm},clip,width=0.208\linewidth,height=0.20\linewidth]{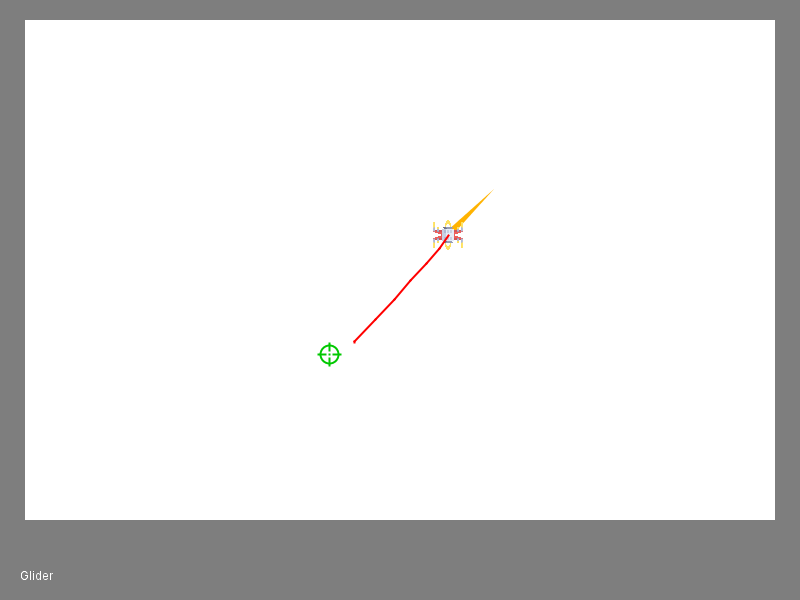}}
	\fbox{\includegraphics[trim={9cm 5.6cm 10cm 6.5cm},clip,width=0.208\linewidth,height=0.20\linewidth]{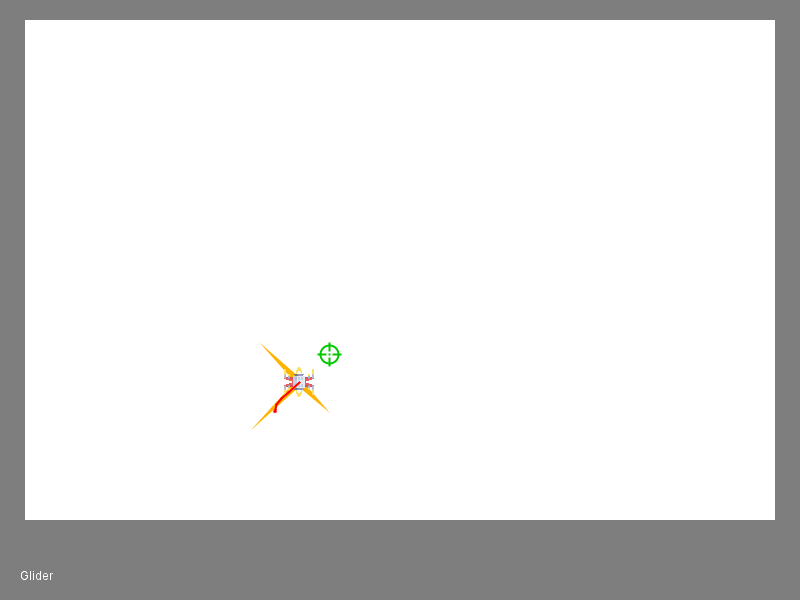}}
	\fbox{\includegraphics[trim={9cm 5.6cm 10cm 6.5cm},clip,width=0.208\linewidth,height=0.20\linewidth]{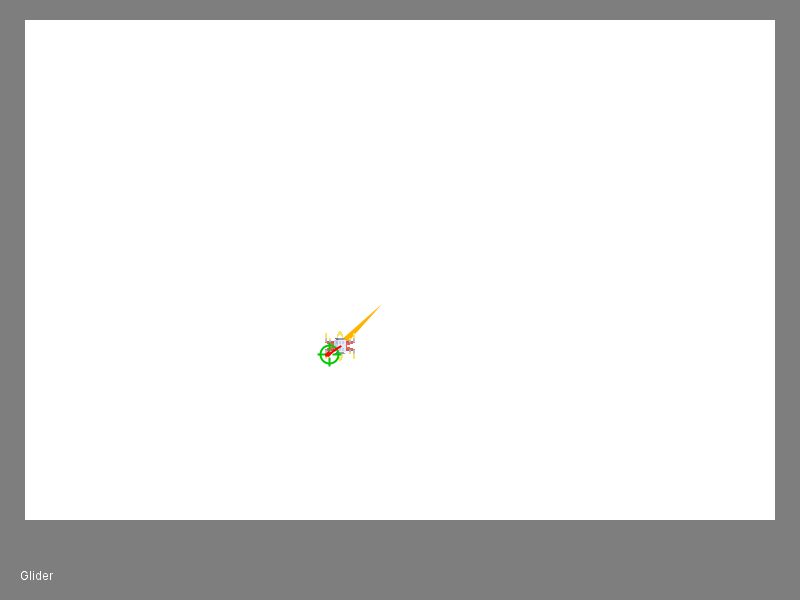}}
	\fbox{\includegraphics[trim={9cm 5.6cm 10cm 6.5cm},clip,width=0.208\linewidth,height=0.20\linewidth]{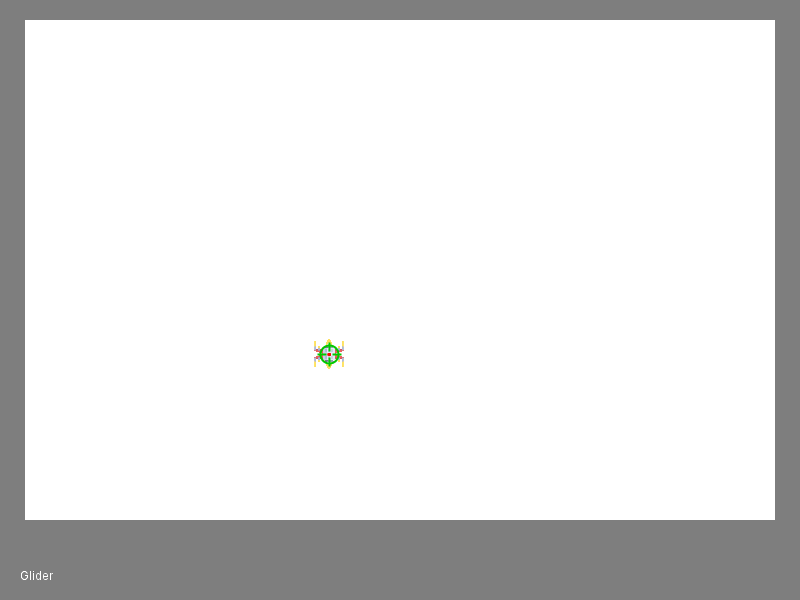}}
	\hfill
	\\
\hfill
	\fbox{\includegraphics[trim={9cm 5.6cm 10cm 6.5cm},clip,width=0.208\linewidth,height=0.20\linewidth]{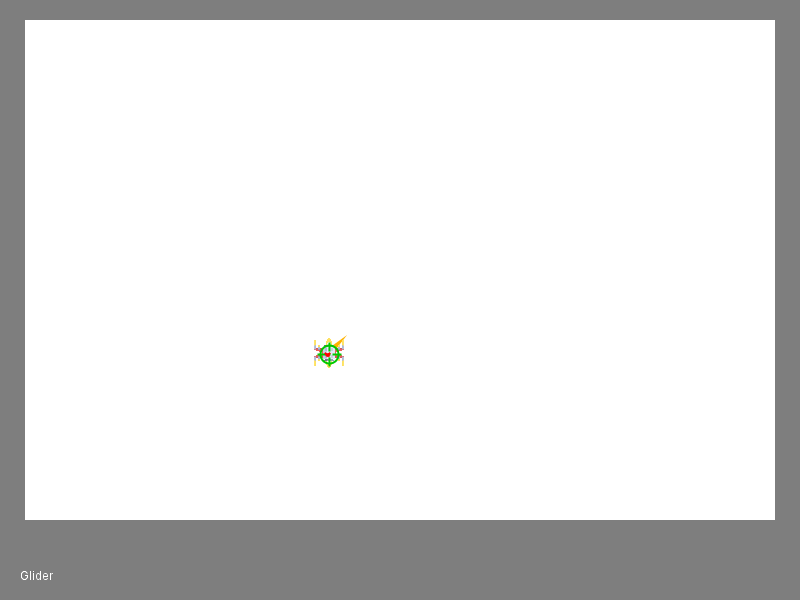}}
	\fbox{\includegraphics[trim={9cm 5.6cm 10cm 6.5cm},clip,width=0.208\linewidth,height=0.20\linewidth]{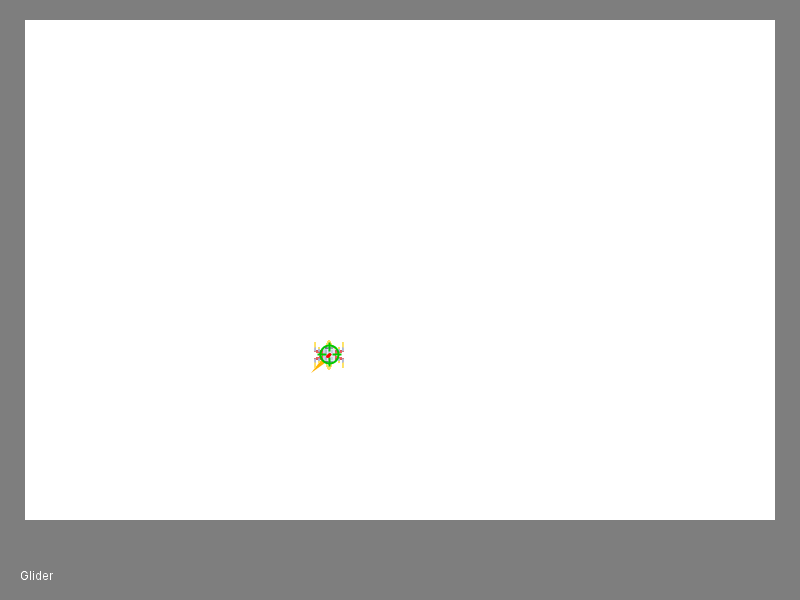}}
	\fbox{\includegraphics[trim={9cm 5.6cm 10cm 6.5cm},clip,width=0.208\linewidth,height=0.20\linewidth]{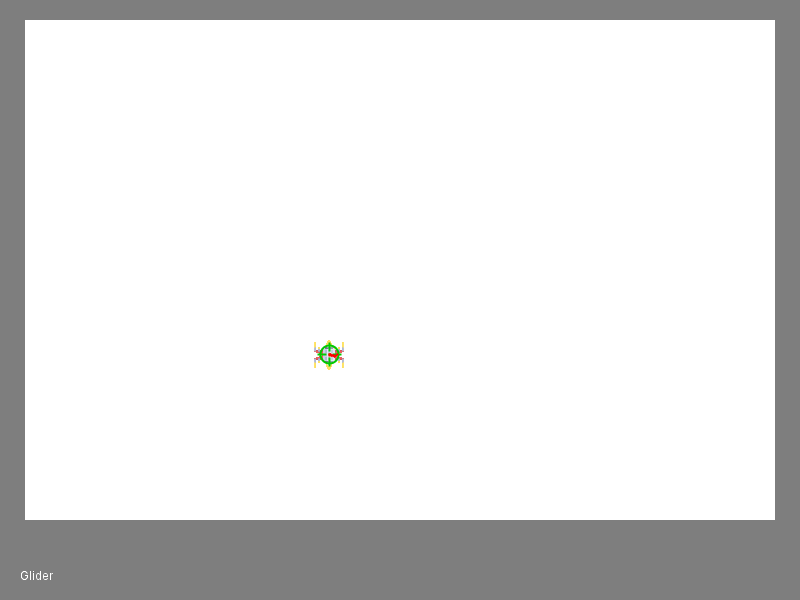}}
	\fbox{\includegraphics[trim={9cm 5.6cm 10cm 6.5cm},clip,width=0.208\linewidth,height=0.20\linewidth]{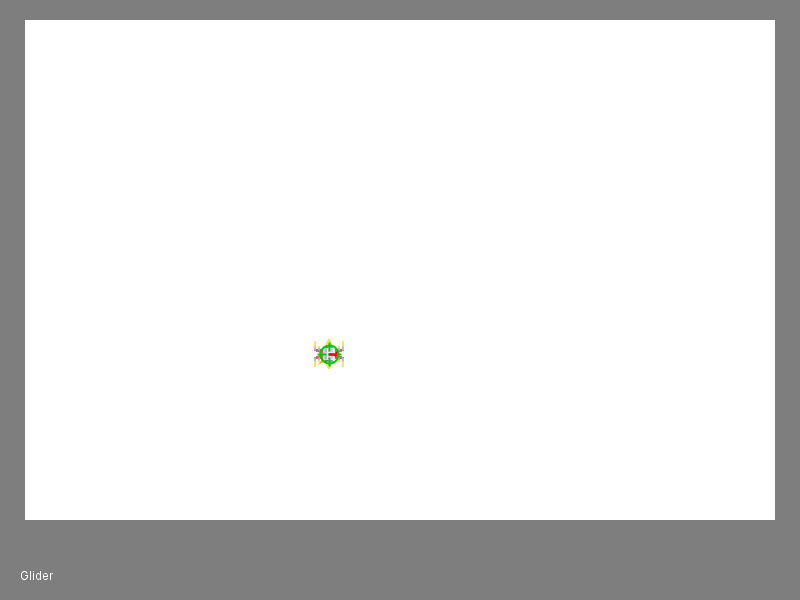}}
	\vspace{.1cm}
		\\
	\fbox{\includegraphics[trim={8.0744cm 8.6cm 7cm 1.0cm},clip,width=0.208\linewidth,height=0.20\linewidth]{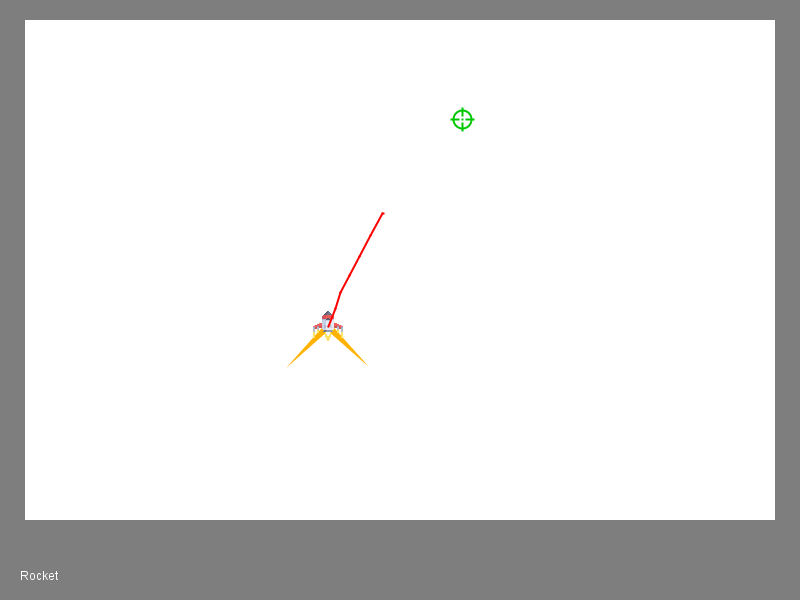}}
	\fbox{\includegraphics[trim={8.0744cm 8.6cm 7cm 1.0cm},clip,width=0.208\linewidth,height=0.20\linewidth]{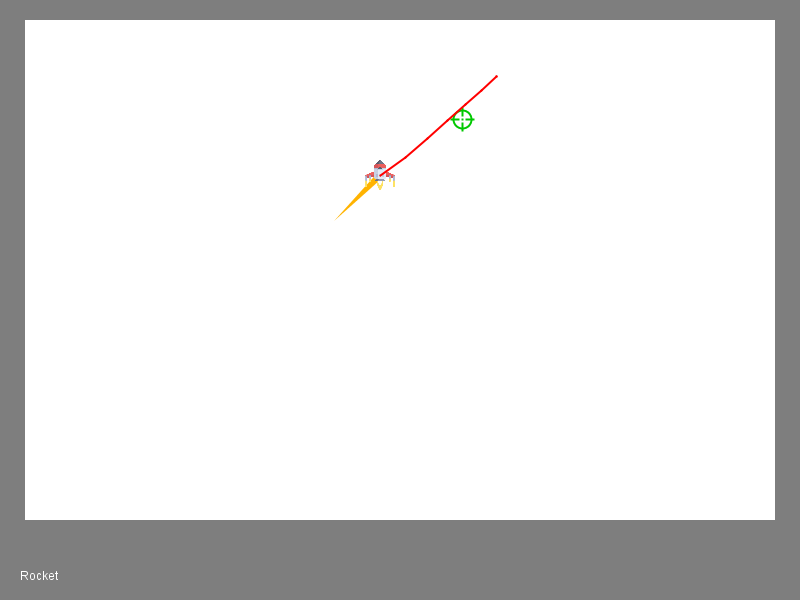}}
	\fbox{\includegraphics[trim={8.0744cm 8.6cm 7cm 1.0cm},clip,width=0.208\linewidth,height=0.20\linewidth]{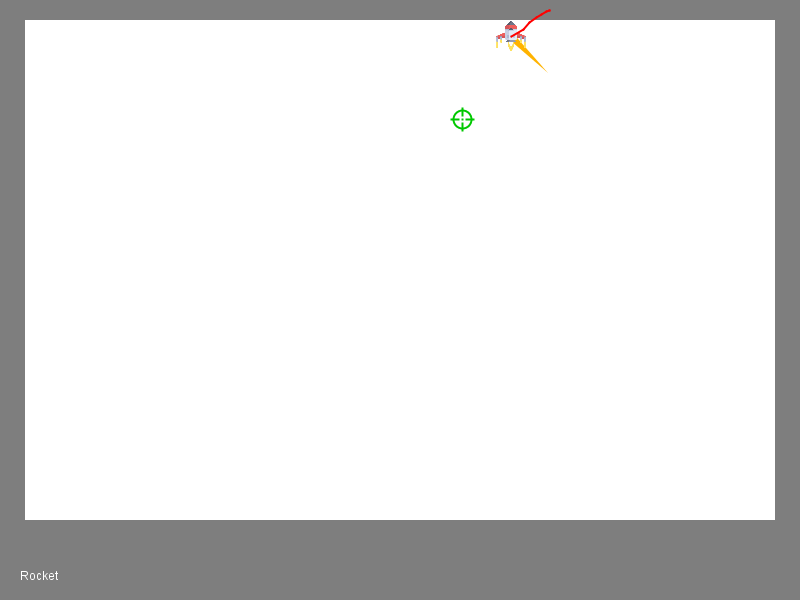}}
	\fbox{\includegraphics[trim={8.0744cm 8.6cm 7cm 1.0cm},clip,width=0.208\linewidth,height=0.20\linewidth]{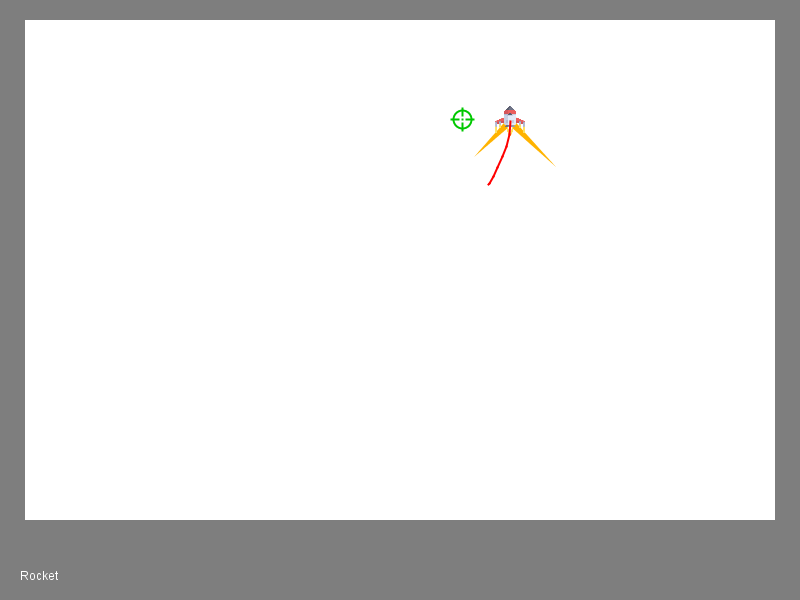}}
	\hfill
	\\
	\hfill
	\fbox{\includegraphics[trim={8.0744cm 8.6cm 7cm 1.0cm},clip,width=0.208\linewidth,height=0.20\linewidth]{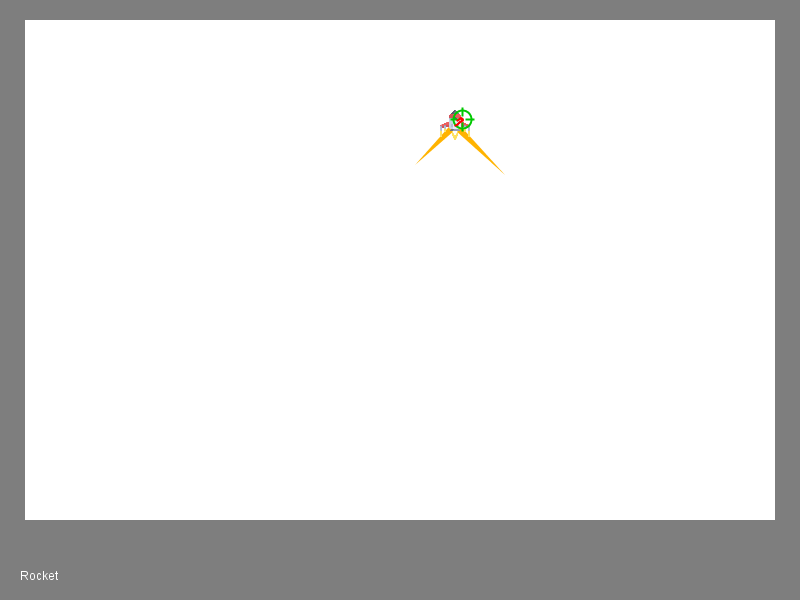}}
	\fbox{\includegraphics[trim={8.0744cm 8.6cm 7cm 1.0cm},clip,width=0.208\linewidth,height=0.20\linewidth]{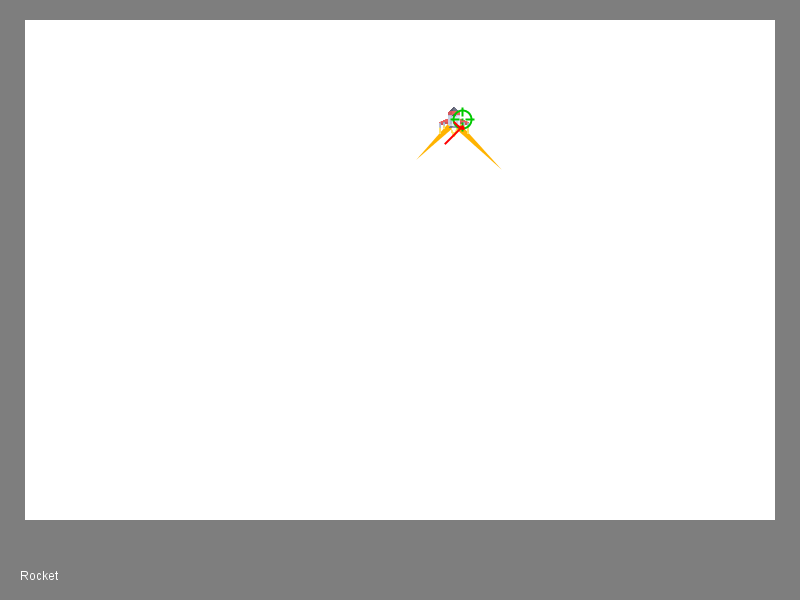}}
	\fbox{\includegraphics[trim={8.0744cm 8.6cm 7cm 1.0cm},clip,width=0.208\linewidth,height=0.20\linewidth]{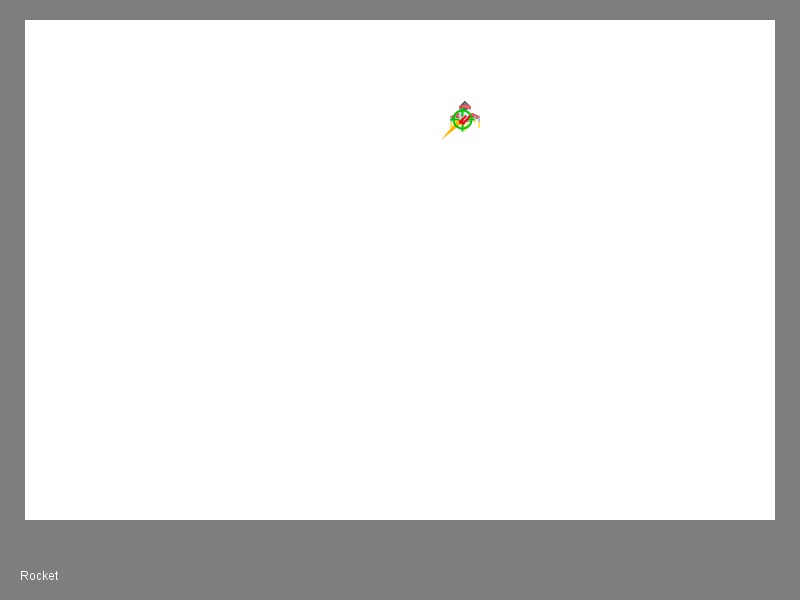}}
	\fbox{\includegraphics[trim={8.0744cm 8.6cm 7cm 1.0cm},clip,width=0.208\linewidth,height=0.20\linewidth]{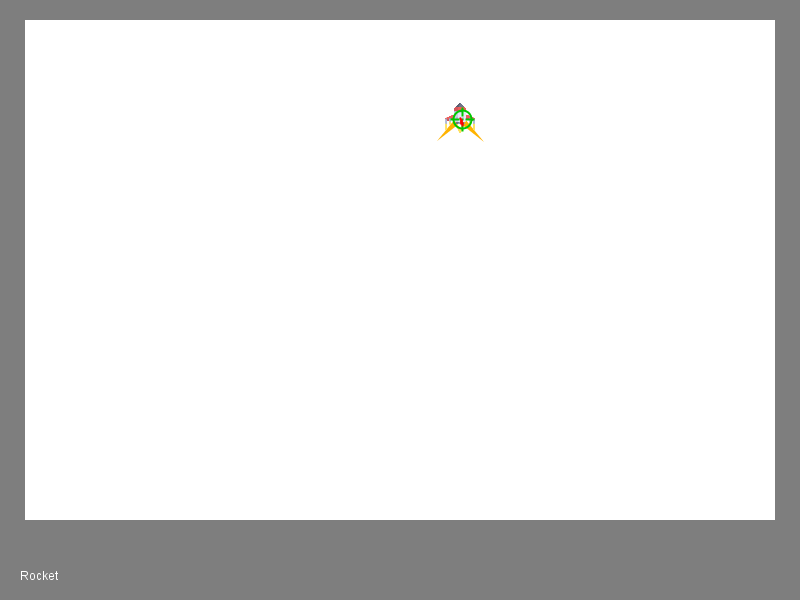}}
	\vspace{.1cm}
		\\
	\fbox{\includegraphics[trim={8.2149cm 7.6cm 10cm 4.0cm},clip,width=0.208\linewidth,height=0.20\linewidth]{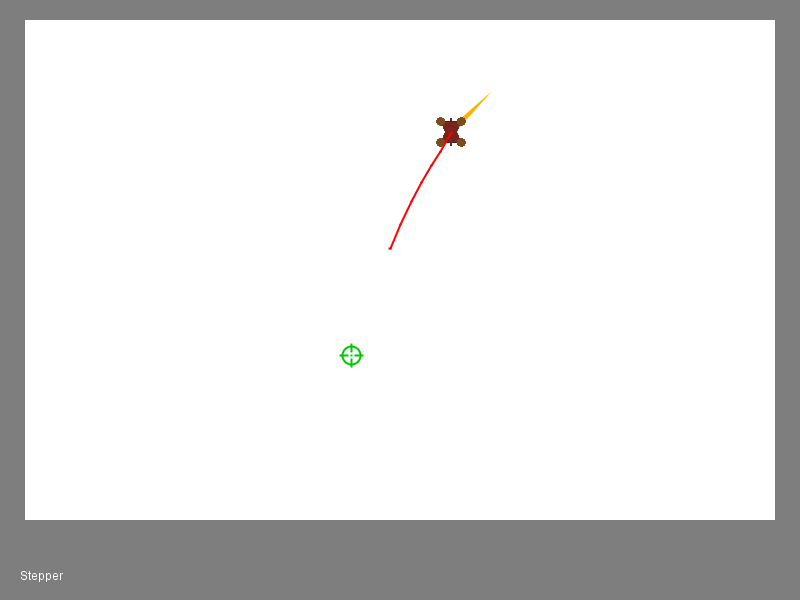}}
	\fbox{\includegraphics[trim={8.2149cm 7.6cm 10cm 4.0cm},clip,width=0.208\linewidth,height=0.20\linewidth]{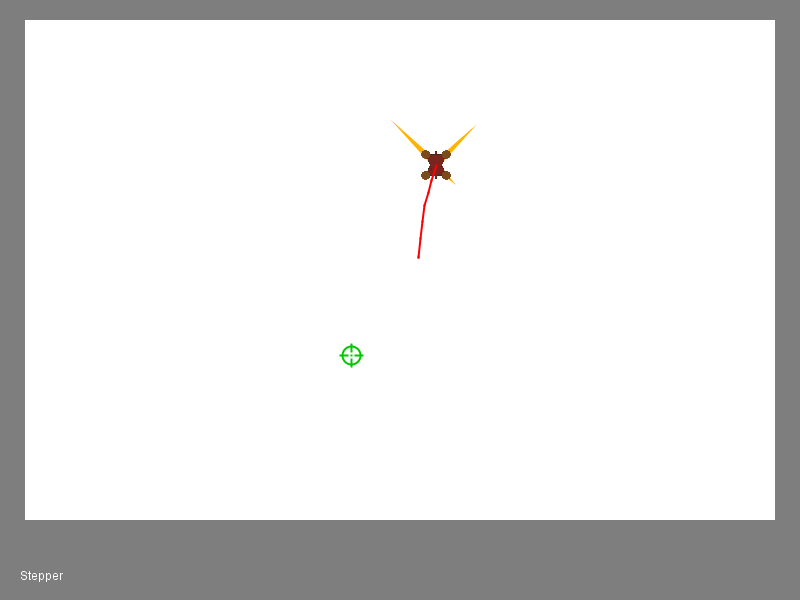}}
	\fbox{\includegraphics[trim={8.2149cm 7.6cm 10cm 4.0cm},clip,width=0.208\linewidth,height=0.20\linewidth]{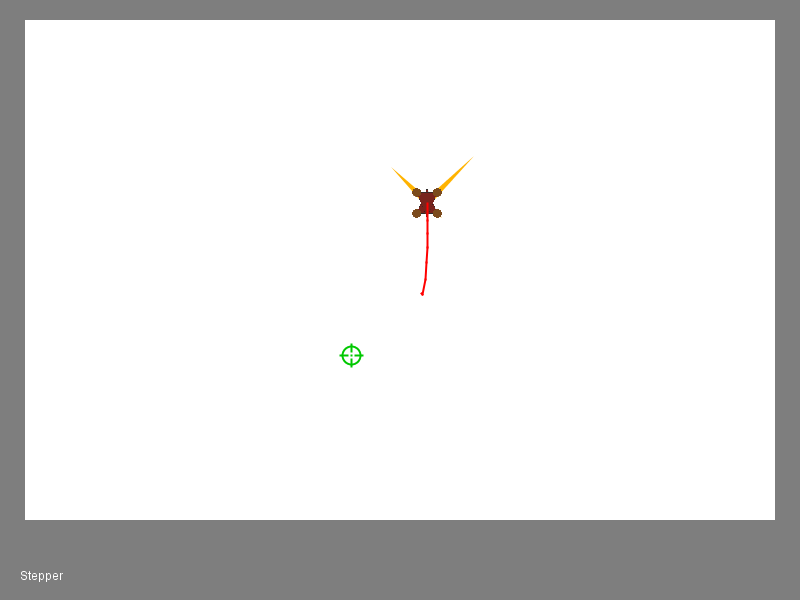}}
	\fbox{\includegraphics[trim={8.2149cm 7.6cm 10cm 4.0cm},clip,width=0.208\linewidth,height=0.20\linewidth]{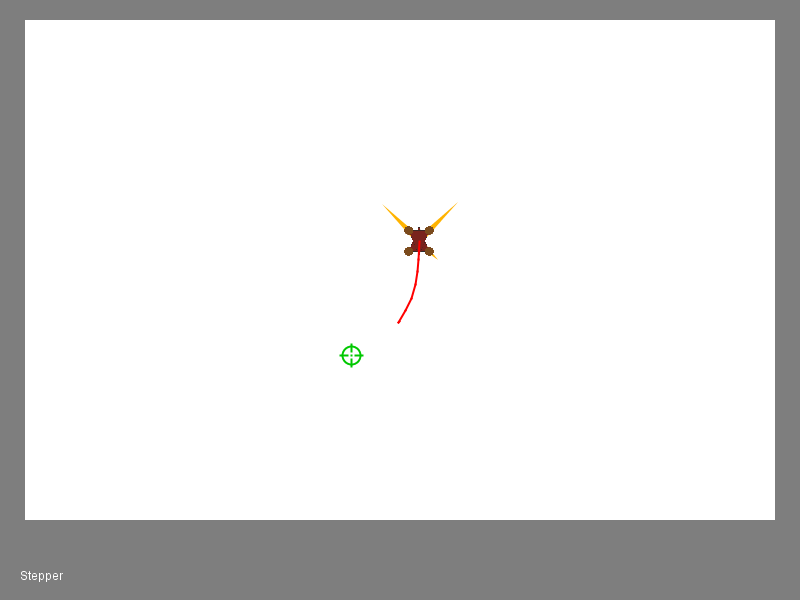}}
	\hfill
	\\
	\hfill
	\fbox{\includegraphics[trim={8.2149cm 7.6cm 10cm 4.0cm},clip,width=0.208\linewidth,height=0.20\linewidth]{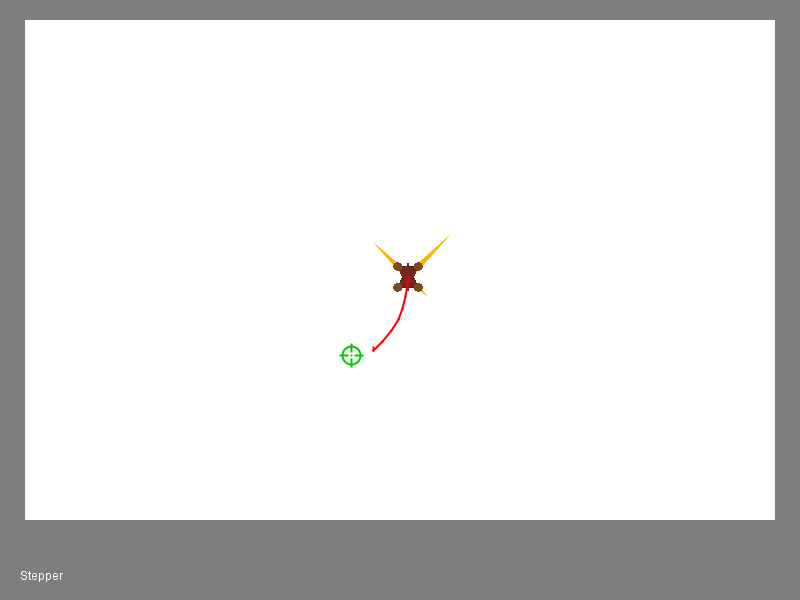}}
	\fbox{\includegraphics[trim={8.2149cm 7.6cm 10cm 4.0cm},clip,width=0.208\linewidth,height=0.20\linewidth]{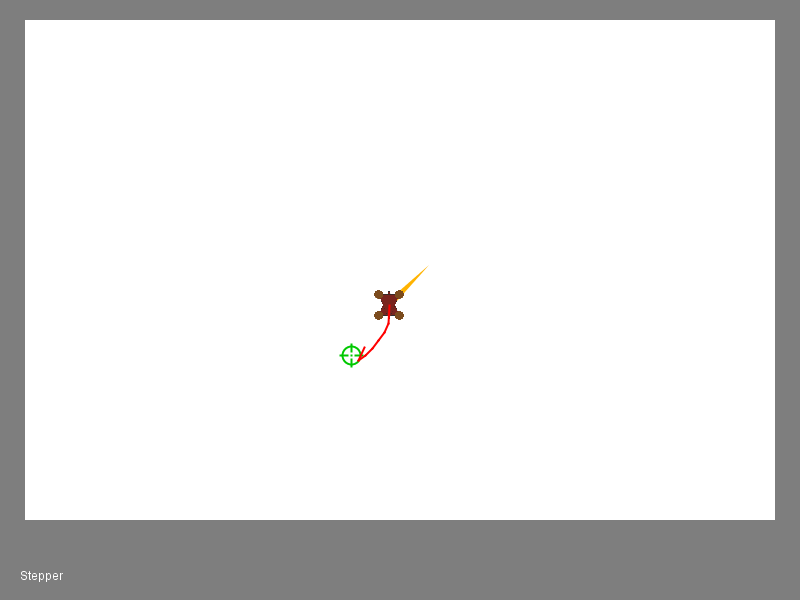}}
	\fbox{\includegraphics[trim={8.2149cm 7.6cm 10cm 4.0cm},clip,width=0.208\linewidth,height=0.20\linewidth]{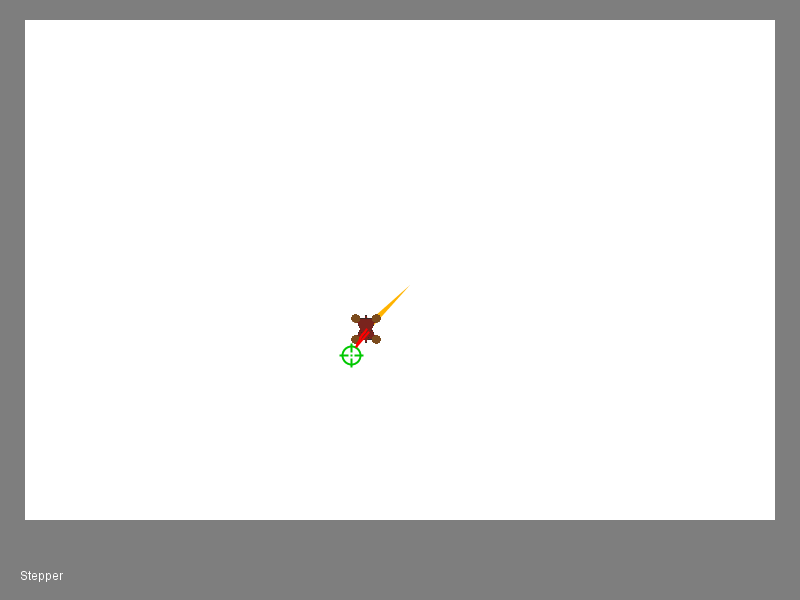}}
	\fbox{\includegraphics[trim={8.2149cm 7.6cm 10cm 4.0cm},clip,width=0.208\linewidth,height=0.20\linewidth]{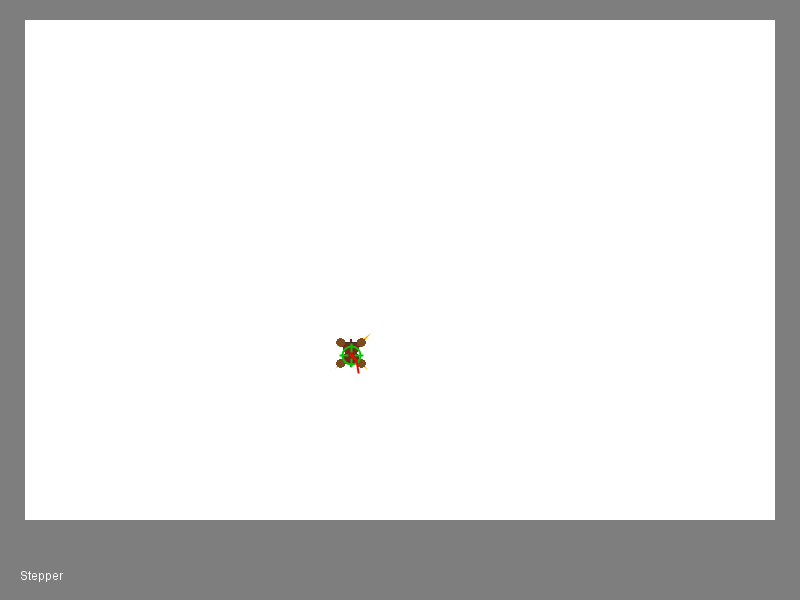}}
	\caption{Typical flight sequence for the three vehicle types controlled by \METHOD*, showing 8 screenshots of glider, rocket, and stepper, which are 10 time steps apart successively in the upper, middle, and bottom two rows, respectively. The green target is approached. The red lines show the current trajectory anticipation of \METHOD*.}    
	\label{figure:flight}
\end{figure*}

\subsection{\METHOD* Performance}
To evaluate the robustness and abilities of \METHOD*, including all relevant settings, we contrast the resulting control performance of the RNN with 36 hidden neurons with the LSTM with 16 hidden units, which have approximately an equal number of weights (1692 versus 1680, respectively). 
Each network was tested to reach a sequence of 50 uniformly randomly positioned targets within a centered inner area of size $1.5\times 1.5$ units, i.e. $\smash{\accentset{\star}{\vec{s}}} \in [-.75,.75] \times [.25,1.75]$ of the rectangular environment.
Thereby, the simulation is divided into a sequence of discrete ‘events’, where the agent ‘becomes’ one of the vehicles $\phi_i$ for $150$ time steps. 
One of the agent’s tasks is to infer which of these events is under way at any given time. 
The values in the tables below are averages over the $20$ independently trained networks and $50$ considered targets, whereby the target positions and vehicle successions were the same for all runs.

We applied Adam in all inference processes. 
Prospective inference was always $P=7$ steps into the future, executing the inference cycle $p=20$ times.
Elsewhere, we have studied this influence (without retrospective inference), showing that even longer prospective inferences are generally possible \citep{Otte:2017}.
Detailed evaluations were run contrasting different learning rates $\eta_c$ and $\eta_\sigma$ for the retrospective context $\vec{c}$ and system state $\vec{\sigma}$ inference. 
In our standard setting, retrospective inference covered $R=20$ time steps into the past, while $r=20$ inference cycles were performed.
In our experience, the retrospective horizon $R$ may vary quite a bit still yielding robust results, while the number of gradient descent cycles needs to be sufficiently large $r>2$, but again a rather wide value range yielded comparable results. 
Note that during optimization the motor commands and the context inputs were clamped to their value range $[0,1]$, and the neural hidden states $\vec{\sigma}$ were clamped in accordance to the range of the respective neurons' activation function.

Figure~\ref{figure:flight} shows typical flight sequences generated by an LSTM controlled by the \METHOD* algorithm, in ten iteration steps. 
Although glider and rocket initially slightly overshoot the target, 
they quickly zoom in.
For the stepper, the projected path is less direct, which is probably partially the case because the goal is simply not directly reachable in seven steps. 
It should be noted that although the images suggest that the motor effort while staying at the goal is minimized, this is not always the case, as there is currently no incentive in the system that stresses motor effort minimization.

Tables~\ref{tab:lstmdist} and \ref{tab:rnndist} show the average distance to the goal location that remained after $150$ time steps, that is, control iterations after target onset.
The first row of results shows the performance when the context bits are set to the correct values (no state inference) while the hidden states of the LSTM are adapted with varying learning rates $\eta_\sigma$. 
Since the information about which vehicle is currently being controlled is provided, a (much simpler) active motor inference problem is solved, yielding robust and accurate goal reaching behavior.
The next four rows show the performance of \METHOD* when the context information, that is, which vehicle is currently being controlled, is not provided. 
Results for different learning rate combinations are shown: $\eta_c$ determines the strength of adapting the contextual neural states $\vec{c}$, while $\eta_\sigma$ controls the hidden neural state $\vec{\sigma}$ adaptations.  
Clearly, overly large or small values yield mediocre performance. 
However, quite a large value range yields robust target reaching behavior. 
Consistently the best setting is with  $\eta_c=.01$ and $\eta_\sigma=.001$, adapting the context bits ten times faster than the hidden states, which is most likely the case because without proper context inputs, overly fast hidden state adaptations will lead to unstable behavior. 
Thus, most robust performance is reached when both context and hidden state activities are adapted, yielding performance that is actually competitive --- in the RNN case even superior --- to the one when the context information is provided!
In sum, with sufficiently small state inference learning rates  $\eta_\sigma$, the additional state inference (much harder problem, no context bit information provided) does not affect performance in a negative manner!

\begin{table}[htb!]
\vspace{-.5cm}
	\centering
	\caption{LSTM: Distance to target after 150 control steps}
	\begin{tabular}{c | ccccc}
		\toprule
		Average & $\eta_\sigma$=0 & $\eta_\sigma$=1e-4 & $\eta_\sigma$=.001 & $\eta_\sigma$=.01 & $\eta_\sigma$=.1\\
		\midrule
		$\vec{c}$ set & 0.006 & 0.006 & 0.006 & 0.007 & 0.051 \\
		$\eta_c$=1e-4 & 0.082 & 0.057 & 0.034 & \emptycell & \emptycell \\
		$\eta_c$=.001 & 0.039 & 0.024 & 0.011 & 0.011 & \emptycell \\
		$\eta_c$=.01 & 0.024 & \emptycell & 0.006 & 0.007 & 0.055 \\
		$\eta_c$=.1 & 0.025 & \emptycell & \emptycell & 0.008 & 0.038 \\ 
		\midrule
		Rocket & $\eta_\sigma$=0 & $\eta_\sigma$=1e-4 & $\eta_\sigma$=.001 & $\eta_\sigma$=.01 & $\eta_\sigma$=.1\\
		\midrule
		$\vec{c}$ set & 0.014 & 0.014 & 0.013 & 0.015 & 0.095 \\
		$\eta_c$=1e-4 & 0.069 & 0.052 & 0.022 & \emptycell & \emptycell \\
		$\eta_c$=.001 & 0.041 & 0.037 & 0.013 & 0.011 & \emptycell \\
		$\eta_c$=.01 & 0.026 & \emptycell & 0.006 & 0.011 & 0.069 \\
		$\eta_c$=.1 & 0.028 & \emptycell & \emptycell & 0.010 & 0.051 \\
		\midrule
		Stepper & $\eta_\sigma$=0 & $\eta_\sigma$=1e-4 & $\eta_\sigma$=.001 & $\eta_\sigma$=.01 & $\eta_\sigma$=.1\\
		\midrule
		$\vec{c}$ set & 0.002 & 0.001 & 0.001 & 0.001 & 0.006 \\
		$\eta_c$=1e-4 & 0.063 & 0.043 & 0.022 & \emptycell & \emptycell \\
		$\eta_c$=.001 & 0.037 & 0.017 & 0.008 & 0.014 & \emptycell \\
		$\eta_c$=.01 & 0.024 & \emptycell & 0.007 & 0.005 & 0.035 \\
		$\eta_c$=.1 & 0.025 & \emptycell & \emptycell & 0.004 & 0.024 \\
		\midrule
		Glider & $\eta_\sigma$=0 & $\eta_\sigma$=1e-4 & $\eta_\sigma$=.001 & $\eta_\sigma$=.01 & $\eta_\sigma$=.1\\
		\midrule
		$\vec{c}$ set & 0.004 & 0.004 & 0.004 & 0.004 & 0.053 \\
		$\eta_c$=1e-4 & 0.116 & 0.078 & 0.060 & \emptycell & \emptycell \\
		$\eta_c$=.001 & 0.040 & 0.017 & 0.012 & 0.009 & \emptycell  \\
		$\eta_c$=.01 & 0.021 & \emptycell & 0.005 & 0.006 & 0.061 \\
		$\eta_c$=.1 & 0.021 &\emptycell  & \emptycell & 0.010 & 0.039 \\
		\bottomrule
	\end{tabular}
\vspace{-.25cm}
	\label{tab:lstmdist}
\end{table}

\begin{table}[htb!]
	\centering
	\caption{RNN: Distance to target after 150 control steps}
	\begin{tabular}{c | c c c c c}
		\toprule
		Average & $\eta_\sigma$=0 & $\eta_\sigma$=1e-4 & $\eta_\sigma$=.001 & $\eta_\sigma$=.01 & $\eta_\sigma$=.1\\
		\midrule
		$\vec{c}$ set & 0.037 & 0.039 & 0.037 & 0.035 & 0.036 \\
		$\eta_c$=1e-4 & 0.162 & 0.019 & 0.019 & \emptycell & \emptycell \\
		$\eta_c$=.001 & 0.038 & 0.052 & 0.020 & 0.018 & \emptycell \\
		$\eta_c$=.01 & 0.040 & \emptycell & 0.017 & 0.019 & 0.029 \\
		$\eta_c$=.1 & 0.169 & \emptycell & \emptycell & 0.044 & 0.034 \\
		\bottomrule
	\end{tabular}
	\label{tab:rnndist}
\end{table}

Table~\ref{tab:lstmdist} additionally shows the performance differences when focusing in on the three different vehicle types in the LSTM case. 
While the parameter dependencies are very similar, the results indicate that it was most difficult to move the rocket towards and keep it close to the goal location.
This is most likely due to the fact that gravity needs to be continuously counteracted in the case of the rocket, but not in the case of the stepper or glider. 


\begin{table}[t]
	\centering
	\caption{LSTM: Average accumulated distance to target}
	\begin{tabular}{c | ccccc}
		\toprule
		Average &  $\eta_\sigma$=0 & $\eta_\sigma$=1e-4 & $\eta_\sigma$=.001 & $\eta_\sigma$=.01 & $\eta_\sigma$=.1\\
		\midrule
		$\vec{c}$ set & 0.061 & 0.060 & 0.060 & 0.061 & 0.109 \\
		$\eta_c$=1e-4 & 0.157 & 0.139 & 0.109 & \emptycell & \emptycell \\
		$\eta_c$=.001 & 0.106 & 0.100 & 0.079 & 0.082 & \emptycell \\
		$\eta_c$=.01 & 0.090 & \emptycell & 0.072 & 0.077 & 0.127 \\
		$\eta_c$=.1 & 0.092 & \emptycell &\emptycell  & 0.077 & 0.115 \\
		\bottomrule
	\end{tabular}
	\label{tab:lstmaccdist}
\end{table}

%

Table~\ref{tab:lstmaccdist} shows for the LSTM case that the average distance to the target object over the 150 steps averaged over all vehicles is smallest when the context information is provided. 
This was indeed the case for all three vehicles (not shown).
This is expectable as context inference inevitably yields erroneous behavior during the first control steps, confirming that the switch in the vehicle identity causes initial disruptions, which are quickly stabilized.

\begin{figure}[t]
	\centering
	\includegraphics[width=\linewidth]{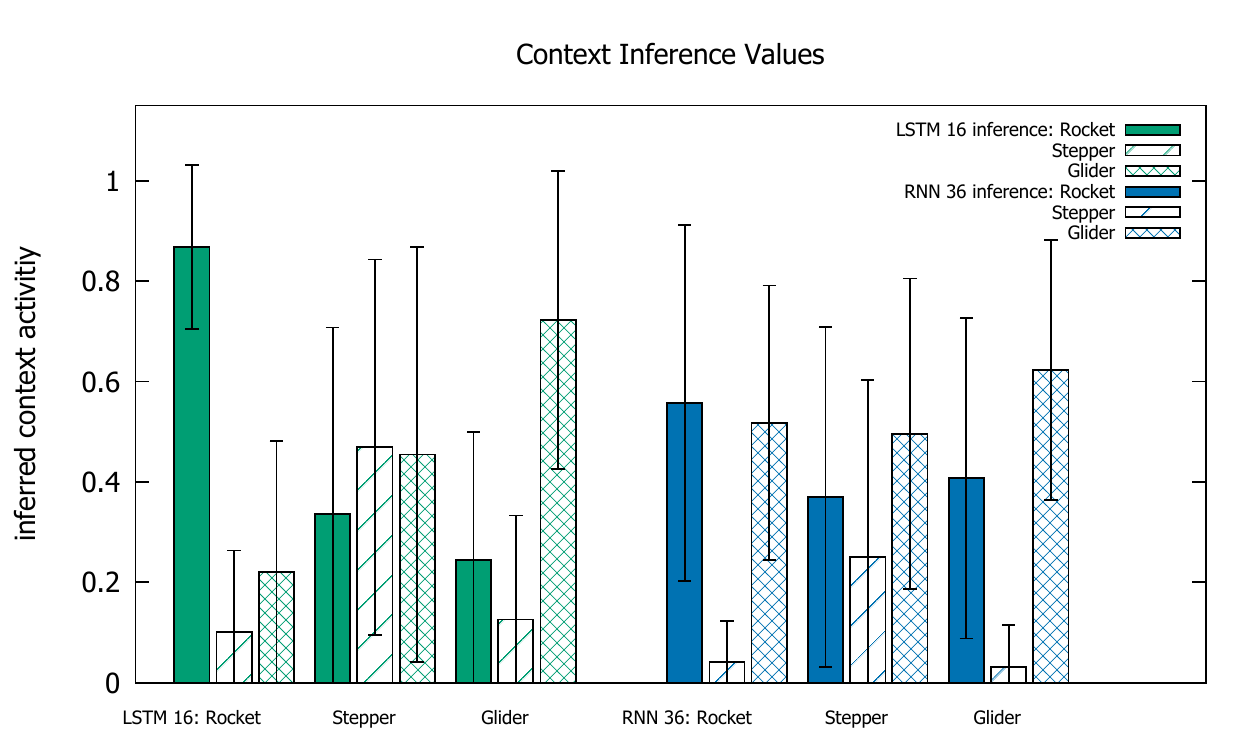}
	\caption{Inferred values of the context values $\vec{c}$ when indicator values were provided during training.}
	\label{figure:ce}
\end{figure}

As a final evaluation, Figure~\ref{figure:ce} shows the inferred context input activations for the three vehicles, contrasting again LSTM with RNN performance. 
The results indicate that the LSTM architecture is better-suited to infer the underlying control system, seeing that the correct vehicle wins in all three cases and the winner is more separated from the two alternatives when contrasted with the RNN performance. 
Clearly, though, the results are very noisy and far from optimal. 
It was observed that once the goal has been reached, the estimates sometimes drifted off towards more incorrect estimates --- probably because the sensorimotor information is not sufficiently informative.
This observation in particular suggests that context estimation stability should be improved by allowing context switches only only when error signals suggest to do so. 
Moreover, active motor inference may be further optimized for the purpose of maintaining high context estimation certainty \citep{Friston:2015}, which should lead to the generation of motor commands that minimize uncertainties in the model state estimates $\vec{\sigma}$.

\section{Emergent Event Encodings}
\label{sec:contextInference}
Despite the apparently rather noisy signals encountered during retrospective context inference above, we were still very interested in whether \METHOD* may be able to infer suitable context vector activities $\vec{c}$ also during model learning.
The experimental setup is identical to the one detailed above, except for the fact that no contextual information is provided at any point, neither about which vehicle is currently being controlled nor about a vehicle switch or about distinct vehicle identities.
As a result, context state $\vec{c}$ inference needs to be applied during learning as well.

\subsection{Control Performance}
Again, we trained ten networks independently (with different weight initializations) on the task.
During training, vehicle switches occurred every $V$ steps (here $V=205$). 
Context vector $\vec{c}$, which we keep at length three, was updated with a rate and depth of $R_c=2$ steps, repeating this retrospective adaptation $r=5$ times with a learning rate of $\eta_c=.1$. 
Weights were updated with a learning rate of $\eta_w=10^{-4}$ and $R_w=30$ steps in depth.
Moreover, every $2000$ steps, all neural activities in the network were reset to zero in order to avoid uncontrolled activity increase.  
During testing, we generated $100$ successive goals (e.g. epochs), which were randomly placed in an inner reachable space of size $1.5 \times 1.5$.
Typically, we admitted $G=150$ steps to reach the goal and switched the vehicle concurrently with the goal switch ($V=150$). 
Figure~\ref{figure:drift} shows the context guesses during a flight sequence controlled by \METHOD*.

When testing active inference-based model-predictive control, the Euclidean distance to the goal locations averaged over all 100 goals and all 10 networks reached a value of $e=.0027$ with an adaptation rate of $\eta_c=.1$. Nearly the same error value was reached when $\eta_c=.01$, while the distance increased to $e=.0057$ and $e=.1406$ with $\eta_c=.001$ and $\eta_c=.0001$, respectively. 
The reached distance is clearly smaller than (i) when context inference is switched off completely, which yields $e=.0998$, (ii) when the learning rate during model learning is lowered to $\eta_c=.01$ yielding $e=.0200$, or (iii) when the vehicle switch occurs more frequently, e.g. every five steps, than the context adaptation, e.g. $R_c=20$, yielding $e=.0429$.  
On the other hand, when the vehicle switch $V$ during learning occurs randomly between every 20th and 30th time step and $R_c=2$, a comparable error of $e=.0032$ is achieved.

Interestingly, as shown above, we achieved a best minimal error of $e=.006$ when the context values were set to one-hot vectors during training, which is larger than $e=.0027$ achieved here.
This implies that the network developed contextual state indicators during learning that are more suitable for the task than the one-hot vectors.
This interpretation is analyzed further in the following section. 
The results imply furthermore that a reasonable wide parameter range yields comparable performance.
Nonetheless, context state adaptations need to occur more frequently than vehicle switches (i.e. event changes) and need to be adapted with a sufficiently large adaptation rate, e.g., $\eta_c=.1$.



%
\subsection{Current Controlled Vehicle Inference}
\begin{figure}[t]
	\centering
	\includegraphics[width=0.6\linewidth]{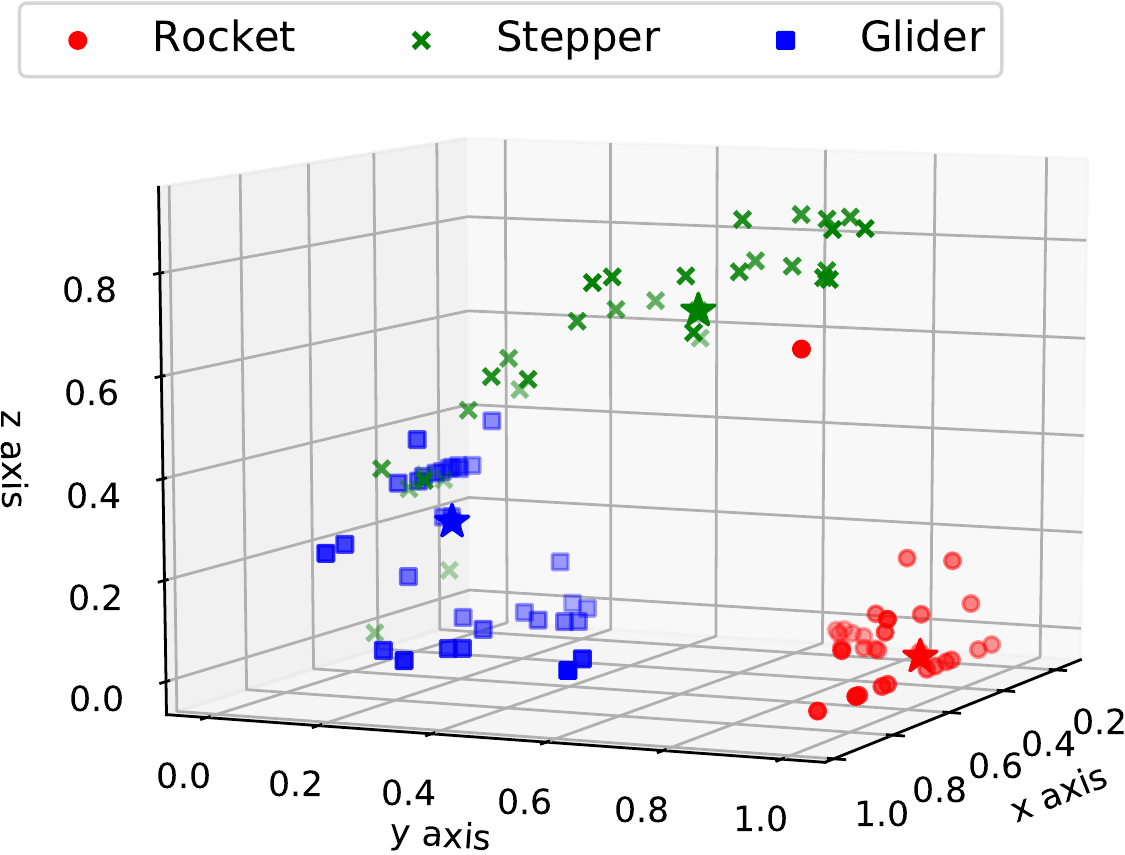}
	\caption{Context guesses plotted when the goal location is reached. The star symbol marks the center of the cluster.}
	\label{figure:clusters}
\end{figure}

Seeing the improved control performance when context inference is switched on during training (i.e. no one-hot vectors), we analyzed the development of the context state estimates during testing further.
Here we focus on reporting the results for one of the ten networks, which serves well for illustrating the main points.
Qualitatively, the results look similar for the other networks.
Figure~\ref{figure:clusters} shows the adapted context values $\vec{c}$ before a goal switch, that is, at the end of each of the 100 epochs, colored according to the vehicle that was just controlled.
The results show three observable clusters, indicating that the network has learned to separate the sensorimotor dynamics of the three vehicles into distinct but somewhat overlapping clusters.

\begin{figure}[t]
	\centering
	\includegraphics[width=0.95\linewidth]{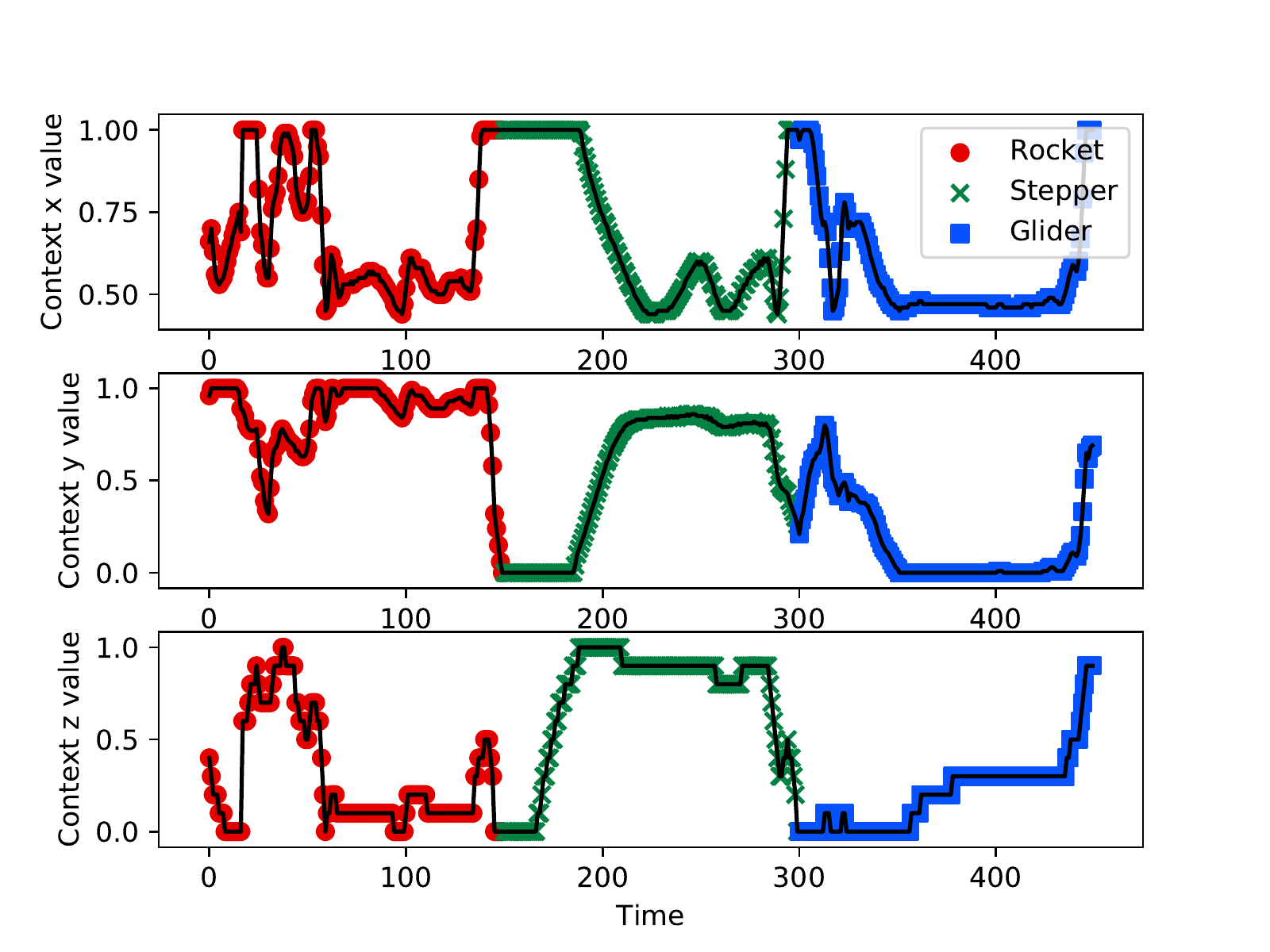}
	\caption{Typical development of the three context guess neural values while controlling (unknowingly) the three vehicles consecutively --- each one for 150 steps. The context is adapted via retrospective inference. The context guesses for the three vehicles are well-separated. However, quite some uncertainty in the estimates is inferable.}
	\label{figure:drift}
\end{figure}

Further analyses revealed that the context state estimates can drift severely once the target is reached. 
This is particularly the case because in the setup Stepper and Glider essentially just need to remain still at the goal state, which they can achieve by sending out zero motor commands but also by sending out equally strong ones to all four motors. 
Similarly, the rocket can mimic gravity by its ineffective downwards thrust motors. 
Thus, particularly at the goal, the sensorimotor signal about which vehicle is currently controlled can become ambiguous. 
Figure~\ref{figure:drift} shows such an exemplary case, while the stepper is controlled.
During the initial steps, the signal from the previous vehicle still influences the gradient. 
Then, the signal tends approximately towards the preferred center of the stepper's context state vector. 
After reaching the goal, however, a drift of the vector can be noticed.

To avoid recording the context guess at the end after it has drifted, we recorded the context guess at the time step when the distance to the goal was minimal.
Figure~\ref{figure:clusters} shows the corresponding results, revealing somewhat focused clustering.
One can notice that the network is separating the context code for the Rocket (in red) well, but sometimes the context guesses get confused between the Stepper (in green) and the Glider (in blue). 
To analyze this observation further, we have calculated the Euclidean distances between the centers of the clusters for the ten independently trained networks
The relative distance between the center of the Rocket cluster and the center of the Glider cluster was the largest on average (d=.415), followed by the distance from the Rocket to the Stepper (d=.348), while the distance between Glider and Stepper was shortest (d=.237).
The gravity effect on the Rocket and the lack of two of the motors seems to require the most distinct encoding. 
On the other hand, the distance between Glider and Stepper is the smallest on average, probably due to the fact that both have 4 motors and do not experience gravity.

\subsection{System Dynamics}
Running the network in a ``context free'' manner did not only raise the question of how the context estimates change over time and where to they converge, but also how these dynamics correlate with goal reaching behavior and the unfolding sensorimotor prediction error dynamics. 
In particular, if further conceptual abstractions of the sensorimotor dynamics are to be fostered, an event boundary signal in the form of a measurable, significant increase in prediction error (akin to \citealp{Gumbsch:2017}) upon vehicle change would be very useful. 

\begin{figure}[t]
	\centering
	\includegraphics[width=\textwidth]{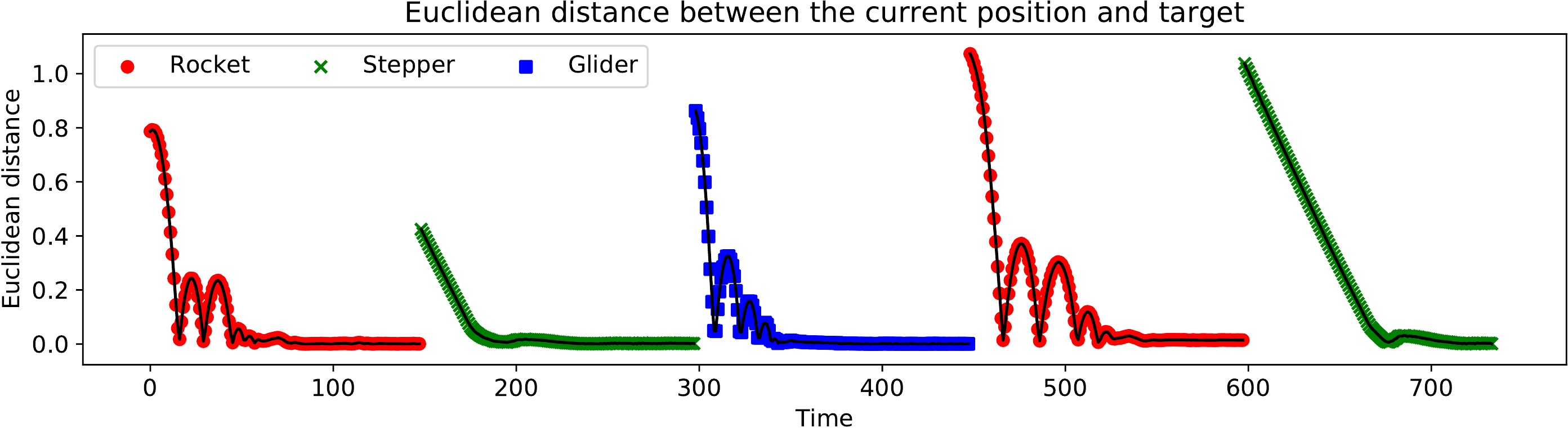}
	\includegraphics[width=\textwidth]{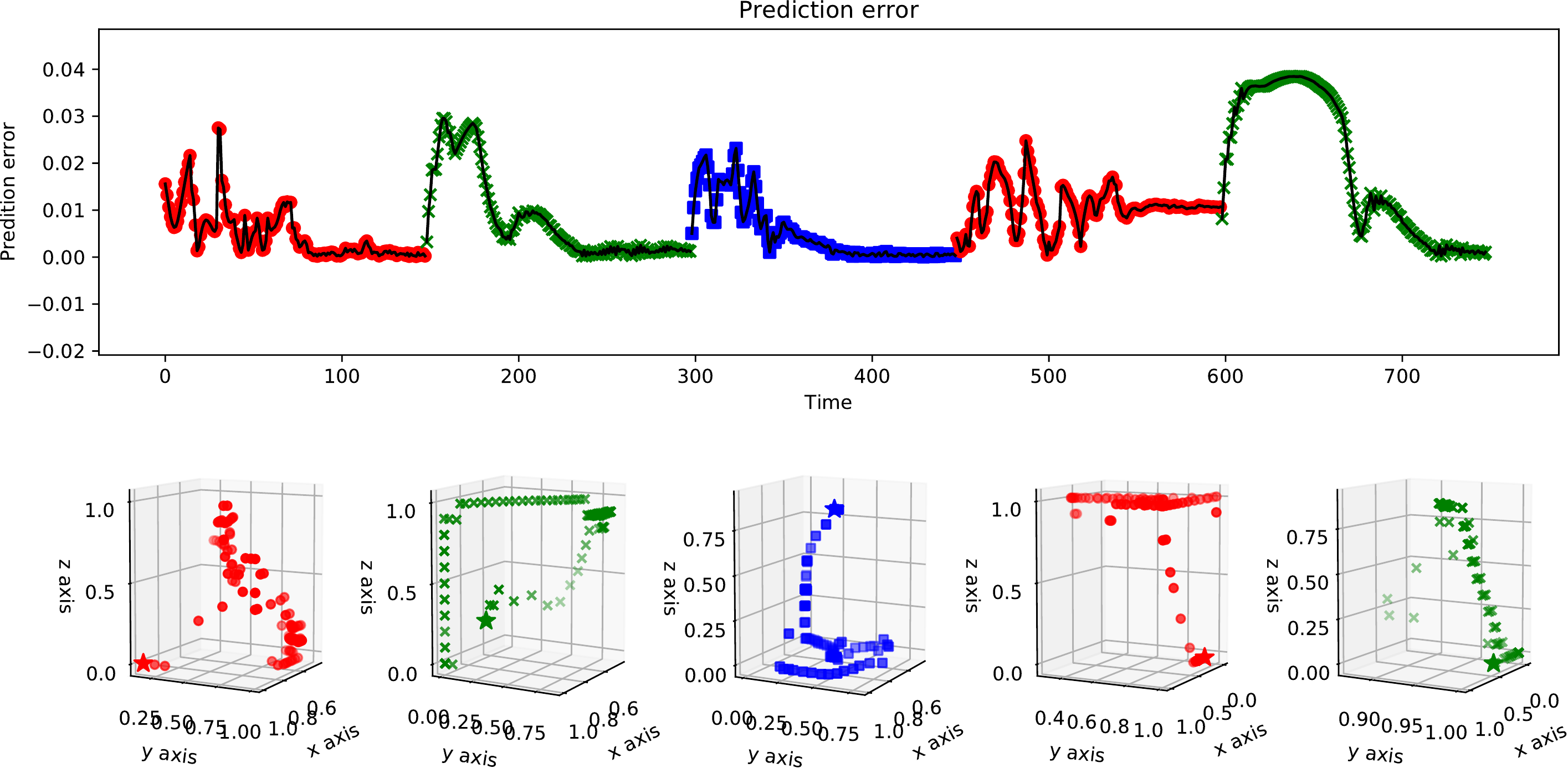}
	%
	\caption{The Euclidean distance between the vehicle and the target at each time step, the prediction error of the network and the changes in the context guess step by step.}  
	\label{figure:PreErr}
\end{figure}
Figure~\ref{figure:PreErr} shows exemplary results for $\approx{750}$ time steps, plotting the Euclidean distance between the vehicle and the target, the current prediction error, as well as the corresponding trajectories of the vehicle estimates, colored and marked by the currently controlled vehicle (unknown to the system).
It is easy to see how the prediction error strongly increases after each vehicle and goal switch ($V,G=150$). 
Moreover, the error decreases when the goal is reached and then stays rather constant at a low but not necessarily 0 level.
The context state estimates furthermore confirm the reported drift behavior. 
Moreover, particularly the second time the rocket is controlled, the context state estimates reach extreme values. 
Gradients that go beyond their boundaries of $[0,1]$ are currently ignored. 
Future versions should consider continuously differentiable context state activation functions, rather than inducing hard boundaries. 

\begin{figure}[t]
	\centering
	\includegraphics[width=\textwidth]{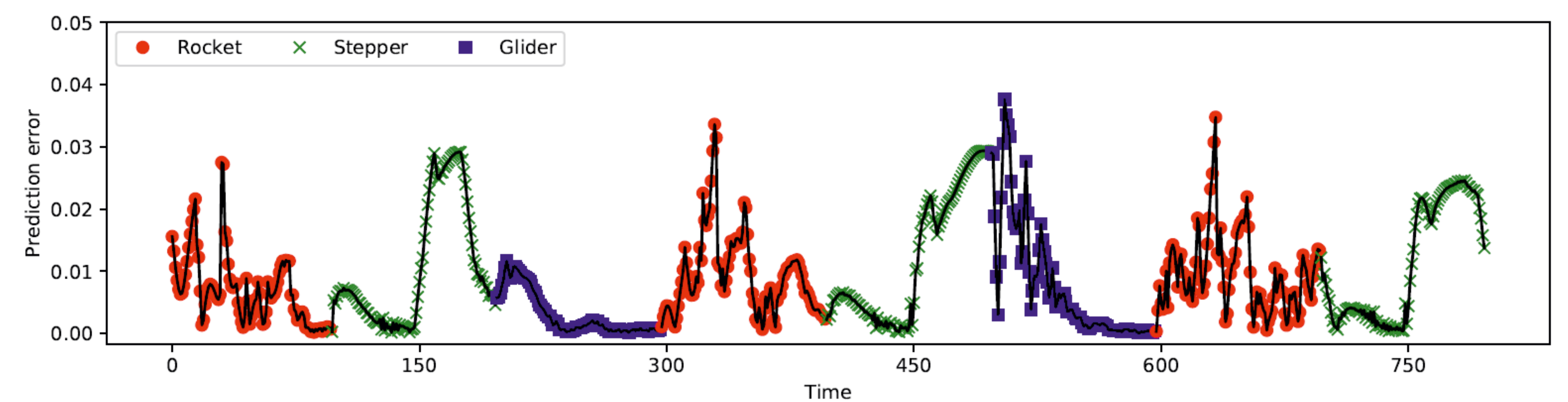}
	%
	\caption{When switching the vehicle asynchronously ($V=100$) to the goal ($G=150$), sudden increases in prediction error still indicate vehicles switches --- albeit less strongly in comparison to combined target and vehicle switches.}  
	\label{figure:PreErrAsy}
\end{figure}

Figure~\ref{figure:PreErrAsy} shows typical prediction error dynamics when vehicle and goal switches ($V=100$, $G=150$) occur asynchronously.
It can be seen that both goal switches and vehicle switches cause temporary increases in prediction error. 
Thus, error dynamics analyses may be used to detect surprise signals, which tend to indicate event boundaries (here vehicle switches) \citep{Butz:2016,Gumbsch:2017}.

%

\section{Summary, Conclusions, and Future Perspectives}
We have shown that \METHOD* maintains ANN activities that 
reflect the past, continuously optimizing its internal (generative) hidden state estimates (both, $\vec{c}^{t-i}$ and $\vec{\sigma}^{t-i}$) about its own body and the environment.
Meanwhile, \METHOD* projects its own state into the future, thus optimizing upcoming environmental interactions (that is, motor commands  $\vec{x}^{t+i}$) under consideration of comparisons between imagined future hidden and actual state estimations (that is, $\vec{\sigma}^{t+i}$, and $\tilde{\vec{s}}^{t+i}$) and desired future goal states (that is, $\smash{\accentset{\star}{\vec{s}}^{t+i}}$). 
We have developed this system as a first step towards sensorimotor-grounded, event-oriented abstractions. 
Essentially, the context vector $\vec{c}$ can be interpreted as signaling the contextual ``event'' the system is currently in. 
We have shown that the event was inferable after model learning but that suitable event encodings can also emerge during learning when inference is applied to both, contextual estimates and model weights.

In particular, the results confirm that \METHOD* is able to identify the currently controlled vehicle without ever being informed about a vehicle switch, the vehicle identities, or the fact that there were three types of vehicles. 
As long as the vehicle type is switched less frequently than the context state estimates are adapted, and as long as the learning rate is sufficiently large, very good control performance was achieved. 
Moreover, distinct contextual encodings developed for each vehicle in a self-organized manner. 
The resulting network structure could be used to minimize prediction error and distance to target, concurrently with deducing the currently controlled vehicle, by performing prospective, goal-directed behavior while retrospectively inferring appropriate contextual states.

Elsewhere, \citet{Butz:2016,Butz:2017a} has proposed that predictive encodings may be suitably compressed into stable event and event boundary codes, which may lead to conceptual abstractions. 
\METHOD* achieves this for the first time in an RNN-based control architecture, inferring contextual neural activities on the fly. 
We see a close relation of these contextual neural activities and stable event codes. 
However, here we have focused on analyzing sensorimotor dynamics, where distinct events are characterized by distinct sensorimotor regimes generated by controlling three different vehicles. 
Similar event encodings may be developed in other scenarios, such as when manipulating objects, using tools, or even when forming place fields for navigating through an environment. 
In all cases, it appears that particular sensorimotor regimes apply---especially when encoding objects in manipulator-relative frames of reference \citep{Calinon:2010,Gumbsch:2017a}. 
To foster the development of such encodings further, unexpected changes in prediction errors should be used as an indicator signal for an event change \citep{Butz:2016,Gumbsch:2017a}.
The gathered results in this respect suggest that these signals are indeed available in \METHOD*.
Once the automatic learning of event encodings is achieved, event-predictive cognition on the compact event-encoding level will become possible, potentially offering a step towards conceptual and compositionally re-combinable event schema abstractions.

It should be kept in mind that the implemented processes currently fully rely on error backpropagation through time. 
Implementations of probabilistic inference processes along similar lines are well-imaginable. 
In addition, training is currently accomplished by providing pseudo-random motor commands. 
Curiosity-driven learning, potentially focusing on expected information gain, is a very attractive alternative, which is indeed also generally compatible with free energy-based active inference \citep{Friston:2009,Oudeyer:2007,Pitti:2017,Schmidhuber:1991}.

Another challenge lies in compressing identified events further when particular trajectories and dynamics become important for achieving particular goals, such as when opening a door or when learning to ride a bicycle.
The addition of event-specific control routine optimization techniques seem to be within the grasp of \METHOD*.
These techniques may focus on achieving particular event transitions as goals and may be implemented by means of policy gradient techniques \citep{Stulp:2013}, 
In this case, but also for improving system scalability and focusing learning in general, the identified event boundary signals may be particularly useful.

Finally, we believe that the exploration of deeper hierarchies, akin to the networks discussed in \citet{Tani:2017}, but with slower, event-adaptive dynamics in deeper levels of the hierarchy, constitutes a highly important next research step.
By focusing RNN learning further on predicting the occurrence of event transitions, it may be possible to develop conceptual abstractions that can be suitably linked with linguistic structures that verbalize executable environmental interactions \citep{Schrodt:2017}.

\subsubsection*{Acknowledgments}
\noindent
Funding from a Feodor Lynen Research Fellowship of the Humboldt Foundation is gratefully acknowledged.

\bibliographystyle{plainnat}
\bibliography{2018-ModeInference_ARXIV_FINAL}

\end{document}

%% file: mydefs.tex
\definecolor{steelblue1}{RGB}{99,184,255}
\definecolor{steelblue2}{RGB}{92,172,238}
\definecolor{steelblue3}{RGB}{79,148,205}
\definecolor{steelblue4}{RGB}{70,130,180}
\definecolor{steelblue5}{RGB}{54,100,139}

\renewcommand{\accentset}[2]{\overset{{}_#1}{#2}{}}
\renewcommand{\vec}[1]{\mathbf{#1}}

\def\netloss{\mathcal{L}}

%% file: 2018-ModeInference_ARXIV_FINAL.bbl
\begin{thebibliography}{48}
\providecommand{\natexlab}[1]{#1}
\providecommand{\url}[1]{\texttt{#1}}
\expandafter\ifx\csname urlstyle\endcsname\relax
  \providecommand{\doi}[1]{doi: #1}\else
  \providecommand{\doi}{doi: \begingroup \urlstyle{rm}\Url}\fi

\bibitem[Arie et~al.(2009)Arie, Endo, Arakaki, Sugano, and Tani]{Arie:2009}
Hiroaki Arie, Tetsuro Endo, Takafumi Arakaki, Shigeki Sugano, and Jun Tani.
\newblock Creating novel goal-directed actions at criticality: A neuro-robotic
  experiment.
\newblock \emph{New Mathematics and Natural Computation}, 05\penalty0
  (01):\penalty0 307--334, 2009.
\newblock \doi{10.1142/S1793005709001283}.

\bibitem[Bar(2009)]{Bar:2009}
Moshe Bar.
\newblock Predictions: {A} universal principle in the operation of the human
  brain.
\newblock \emph{Philosophical Transactions of the Royal Society B: Biological
  Sciences}, 364\penalty0 (1521):\penalty0 1181--1182, 2009.
\newblock \doi{10.1098/rstb.2008.0321}.

\bibitem[Botvinick and Weinstein(2014)]{Botvinick:2014}
Matthew Botvinick and Ari Weinstein.
\newblock Model-based hierarchical reinforcement learning and human action
  control.
\newblock \emph{Philosophical Transactions of the Royal Society of London B:
  Biological Sciences}, 369\penalty0 (1655), 2014.
\newblock ISSN 0962-8436.
\newblock \doi{10.1098/rstb.2013.0480}.

\bibitem[Botvinick et~al.(2009)Botvinick, Niv, and Barto]{Botvinick:2009}
Matthew Botvinick, Yael Niv, and Andrew~C. Barto.
\newblock Hierarchically organized behavior and its neural foundations: A
  reinforcement learning perspective.
\newblock \emph{Cognition}, 113\penalty0 (3):\penalty0 262 -- 280, 2009.
\newblock ISSN 0010-0277.
\newblock \doi{10.1016/j.cognition.2008.08.011}.

\bibitem[Buckner and Carroll(2007)]{Buckner:2007}
Randy~L. Buckner and Daniel~C. Carroll.
\newblock Self-projection and the brain.
\newblock \emph{Trends in Cognitive Sciences}, 11:\penalty0 49--57, 2007.

\bibitem[Butz(2016)]{Butz:2016}
Martin~V. Butz.
\newblock Towards a unified sub-symbolic computational theory of cognition.
\newblock \emph{Frontiers in Psychology}, 7\penalty0 (925), 2016.
\newblock ISSN 1664-1078.
\newblock \doi{10.3389/fpsyg.2016.00925}.

\bibitem[Butz(2017)]{Butz:2017a}
Martin~V. Butz.
\newblock Which structures are out there? learning predictive compositional
  concepts based on social sensorimotor explorations.
\newblock MIND Group, Frankfurt am Main, 2017.
\newblock \doi{10.15502/9783958573093}.

\bibitem[Butz and Kutter(2017)]{Butz:2017}
Martin~V. Butz and Esther~F. Kutter.
\newblock \emph{How the Mind Comes Into Being: Introducing Cognitive Science
  from a Functional and Computational Perspective}.
\newblock Oxford University Press, Oxford, UK, 2017.

\bibitem[Calinon et~al.(2010)Calinon, D'halluin, Sauser, Caldwell, and
  Billard]{Calinon:2010}
Sylvain Calinon, Florent D'halluin, Eric~L. Sauser, Darwin~G. Caldwell, and
  Aude~G. Billard.
\newblock Learning and reproduction of gestures by imitation.
\newblock \emph{Robotics Automation Magazine, IEEE}, 17\penalty0 (2):\penalty0
  44--54, 2010.
\newblock ISSN 1070-9932.
\newblock \doi{10.1109/MRA.2010.936947}.

\bibitem[Camacho and Bordons(1999)]{Camacho:1999a}
E.~F. Camacho and C.~Bordons.
\newblock \emph{Model Predictive Control}.
\newblock Springer-Verlag, 1999.
\newblock ISBN 3-540-76241-8.

\bibitem[Clark(2016)]{Clark:2016}
Andy Clark.
\newblock \emph{Surfing Uncertainty: Prediction, action and the embodied mind}.
\newblock Oxford University Press, Oxford, UK, 2016.

\bibitem[Friston(2009)]{Friston:2009}
Karl Friston.
\newblock The free-energy principle: a rough guide to the brain?
\newblock \emph{Trends in Cognitive Sciences}, 13\penalty0 (7):\penalty0 293 --
  301, 2009.
\newblock ISSN 1364-6613.
\newblock \doi{10.1016/j.tics.2009.04.005}.

\bibitem[Friston et~al.(2015)Friston, Rigoli, Ognibene, Mathys, FitzGerald, and
  Pezzulo]{Friston:2015}
Karl Friston, Francesco Rigoli, Dimitri Ognibene, Christoph Mathys, Thomas
  FitzGerald, and Giovanni Pezzulo.
\newblock Active inference and epistemic value.
\newblock \emph{Cognitive Neuroscience}, 6:\penalty0 187--214, 2015.
\newblock \doi{10.1080/17588928.2015.1020053}.

\bibitem[Gers et~al.(2002)Gers, Schraudolph, and Schmidhuber]{Gers:2002}
Felix~A. Gers, Nicol~N. Schraudolph, and J{\"u}rgen Schmidhuber.
\newblock Learning precise timing with {LSTM} recurrent networks.
\newblock \emph{Journal of Machine Learning Research}, 3:\penalty0 115--143,
  2002.

\bibitem[Gumbsch et~al.(2017{\natexlab{a}})Gumbsch, Otte, and
  Butz]{Gumbsch:2017}
Christian Gumbsch, Sebastian Otte, and Martin~V. Butz.
\newblock A computational model for the dynamical learning of event taxonomies.
\newblock In \emph{Proceedings of the 39th Annual Meeting of the Cognitive
  Science Society}, pages 452--457. Cognitive Science Society,
  2017{\natexlab{a}}.

\bibitem[Gumbsch et~al.(2017{\natexlab{b}})Gumbsch, Otte, and
  Butz]{Gumbsch:2017a}
Christian Gumbsch, Sebastian Otte, and Martin~V. Butz.
\newblock A computational model for the dynamical learning of event taxonomies.
\newblock \emph{Proceedings of the 39th Annual Meeting of the Cognitive Science
  Society}, pages 452--457, 2017{\natexlab{b}}.

\bibitem[Hohwy(2013)]{Hohwy:2013}
Jakob Hohwy.
\newblock \emph{The Predictive Mind}.
\newblock Oxford University Press, Oxford, UK, 2013.

\bibitem[Hommel et~al.(2001)Hommel, M{\"u}sseler, Aschersleben, and
  Prinz]{Hommel:2001}
Bernhard Hommel, Jochen M{\"u}sseler, Gisa Aschersleben, and Wolfgang Prinz.
\newblock The theory of event coding ({TEC}): {A} framework for perception and
  action planning.
\newblock \emph{Behavioral and Brain Sciences}, 24:\penalty0 849--878, 2001.

\bibitem[Kingma and Ba(2014)]{Kingma:2014}
Diederik~P. Kingma and Jimmy~L. Ba.
\newblock Adam: {A} method for stochastic optimization.
\newblock \emph{ArXiv e-prints}, abs/1412.6980, 2014.

\bibitem[Kirkpatrick et~al.(2016)Kirkpatrick, Pascanu, Rabinowitz, Veness,
  Desjardins, Rusu, Milan, Quan, Ramalho, Grabska{-}Barwinska, Hassabis,
  Clopath, Kumaran, and Hadsell]{Kirkpatrick:2016}
James Kirkpatrick, Razvan Pascanu, Neil~C. Rabinowitz, Joel Veness, Guillaume
  Desjardins, Andrei~A. Rusu, Kieran Milan, John Quan, Tiago Ramalho, Agnieszka
  Grabska{-}Barwinska, Demis Hassabis, Claudia Clopath, Dharshan Kumaran, and
  Raia Hadsell.
\newblock Overcoming catastrophic forgetting in neural networks.
\newblock \emph{CoRR}, abs/1612.00796, 2016.

\bibitem[Lake et~al.(2017)Lake, Ullman, Tenenbaum, and Gershman]{Lake:2017}
Brenden~M. Lake, Tomer~D. Ullman, Joshua~B. Tenenbaum, and Samuel~J. Gershman.
\newblock Building machines that learn and think like people.
\newblock \emph{Behavioral and Brain Sciences}, 2017.
\newblock \doi{10.1017/S0140525X16001837}.

\bibitem[McClelland et~al.(2010)McClelland, Botvinick, Noelle, Plaut, Rogers,
  Seidenberg, and Smith]{McClelland:2010}
James~L. McClelland, Matthew~M. Botvinick, David~C. Noelle, David~C. Plaut,
  Timothy~T. Rogers, Mark~S. Seidenberg, and Linda~B. Smith.
\newblock Letting structure emerge: connectionist and dynamical systems
  approaches to cognition.
\newblock \emph{Trends in Cognitive Sciences}, 14\penalty0 (8):\penalty0
  348--356, 2010.
\newblock ISSN 1364-6613.
\newblock \doi{10.1016/j.tics.2010.06.002}.

\bibitem[Mnih et~al.(2015)Mnih, Kavukcuoglu, Silver, Rusu, Veness, Bellemare,
  Graves, Riedmiller, Fidjeland, Ostrovski, Petersen, Beattie, Sadik,
  Antonoglou, King, Kumaran, Wierstra, Legg, and Hassabis]{Mnih:2015}
Volodymyr Mnih, Koray Kavukcuoglu, David Silver, Andrei~A. Rusu, Joel Veness,
  Marc~G. Bellemare, Alex Graves, Martin Riedmiller, Andreas~K. Fidjeland,
  Georg Ostrovski, Stig Petersen, Charles Beattie, Amir Sadik, Ioannis
  Antonoglou, Helen King, Dharshan Kumaran, Daan Wierstra, Shane Legg, and
  Demis Hassabis.
\newblock Human-level control through deep reinforcement learning.
\newblock \emph{Nature}, 518\penalty0 (7540):\penalty0 529--533, February 2015.
\newblock ISSN 0028-0836.
\newblock \doi{10.1038/nature14236}.

\bibitem[{Murata} et~al.(2017){Murata}, {Yamashita}, {Arie}, {Ogata}, {Sugano},
  and {Tani}]{Murata:2017}
S.~{Murata}, Y.~{Yamashita}, H.~{Arie}, T.~{Ogata}, S.~{Sugano}, and J.~{Tani}.
\newblock Learning to perceive the world as probabilistic or deterministic via
  interaction with others: A neuro-robotics experiment.
\newblock \emph{IEEE Transactions on Neural Networks and Learning Systems},
  28\penalty0 (4):\penalty0 830--848, April 2017.
\newblock ISSN 2162-237X.
\newblock \doi{10.1109/TNNLS.2015.2492140}.

\bibitem[Najnin and Banerjee(2017)]{Najnin:2017}
Shamima Najnin and Bonny Banerjee.
\newblock A predictive coding framework for a developmental agent: Speech motor
  skill acquisition and speech production.
\newblock \emph{Speech Communication}, 92:\penalty0 24--41, September 2017.
\newblock ISSN 0167-6393.
\newblock \doi{10.1016/j.specom.2017.05.002}.

\bibitem[Otte et~al.(2017{\natexlab{a}})Otte, Schmitt, Friston, and
  Butz]{Otte:2017}
Sebastian Otte, Theresa Schmitt, Karl Friston, and Martin~V. Butz.
\newblock Inferring adaptive goal-directed behavior within recurrent neural
  networks.
\newblock \emph{26th International Conference on Artificial Neural Networks
  (ICANN17)}, pages 227--235, 2017{\natexlab{a}}.

\bibitem[Otte et~al.(2017{\natexlab{b}})Otte, Zwiener, and Butz]{Otte:2017a}
Sebastian Otte, Adrian Zwiener, and Martin~V. Butz.
\newblock Inherently constraint-aware control of many-joint robot arms with
  inverse recurrent models.
\newblock \emph{26th International Conference on Artificial Neural Networks
  (ICANN17)}, pages 262--270, 2017{\natexlab{b}}.

\bibitem[Oudeyer et~al.(2007)Oudeyer, Kaplan, and Hafner]{Oudeyer:2007}
P.-Y. Oudeyer, F.~Kaplan, and V.~V. Hafner.
\newblock Intrinsic motivation systems for autonomous mental development.
\newblock \emph{IEEE Transactions on Evolutionary Computation}, 11:\penalty0
  265--286, 2007.
\newblock ISSN 1089-778X.
\newblock \doi{10.1109/TEVC.2006.890271}.

\bibitem[Pitti et~al.(2017)Pitti, Gaussier, and Quoy]{Pitti:2017}
Alexandre Pitti, Philippe Gaussier, and Mathias Quoy.
\newblock Iterative free-energy optimization for recurrent neural networks
  (inferno).
\newblock \emph{PLOS ONE}, 12\penalty0 (3):\penalty0 1--33, 03 2017.
\newblock \doi{10.1371/journal.pone.0173684}.

\bibitem[Radvansky and Zacks(2014)]{Radvansky:2014}
Gabriel~A. Radvansky and Jeffrey~M. Zacks.
\newblock \emph{Event cognition}.
\newblock Oxford University Press, Oxford, UK, 2014.

\bibitem[Rao and Ballard(1999)]{Rao:1999}
Rajesh P.~N. Rao and Dana~H. Ballard.
\newblock Predictive coding in the visual cortex: a functional interpretation
  of some extra-classical receptive-field effects.
\newblock \emph{Nature Neuroscience}, 2\penalty0 (1):\penalty0 79--87, January
  1999.
\newblock \doi{10.1038/4580}.

\bibitem[Richmond et~al.(2017)Richmond, Gold, and Zacks]{Richmond:2017}
Lauren~L. Richmond, David~A. Gold, and Jeffrey~M. Zacks.
\newblock Event perception: Translations and applications.
\newblock 6\penalty0 (2):\penalty0 111--120, 2017.
\newblock ISSN 2211-3681.
\newblock \doi{10.1016/j.jarmac.2016.11.002}.

\bibitem[Schacter et~al.(2012)Schacter, Addis, Hassabis, Martin, Spreng, and
  Szpunar]{Schacter:2012}
Daniel~L. Schacter, Donna~Rose Addis, Demis Hassabis, Victoria~C. Martin,
  R.~Nathan Spreng, and Karl~K. Szpunar.
\newblock The future of memory: Remembering, imagining, and the brain.
\newblock \emph{Neuron}, 76\penalty0 (4):\penalty0 677--694, November 2012.
\newblock ISSN 0896-6273.
\newblock \doi{10.1016/j.neuron.2012.11.001}.

\bibitem[Schmidhuber(1991)]{Schmidhuber:1991}
Jurgen Schmidhuber.
\newblock A possibility for implementing curiosity and boredom in
  model-building neural controllers.
\newblock \emph{Proceedings of the first international conference on simulation
  of adaptive behavior: From animals to animats}, pages 222--227, 1991.

\bibitem[Schrodt et~al.(2017)Schrodt, Kneissler, Ehrenfeld, and
  Butz]{Schrodt:2017}
Fabian Schrodt, Jan Kneissler, Stephan Ehrenfeld, and Martin~V. Butz.
\newblock Mario becomes cognitive.
\newblock \emph{Topics in Cognitive Science}, 9\penalty0 (2):\penalty0
  343–373, 2017.
\newblock \doi{10.1111/tops.12252}.

\bibitem[Silver et~al.(2016)Silver, Huang, Maddison, Guez, Sifre, van~den
  Driessche, Schrittwieser, Antonoglou, Panneershelvam, Lanctot, Dieleman,
  Grewe, Nham, Kalchbrenner, Sutskever, Lillicrap, Leach, Kavukcuoglu, Graepel,
  and Hassabis]{Silver:2016}
David Silver, Aja Huang, Chris~J. Maddison, Arthur Guez, Laurent Sifre, George
  van~den Driessche, Julian Schrittwieser, Ioannis Antonoglou, Veda
  Panneershelvam, Marc Lanctot, Sander Dieleman, Dominik Grewe, John Nham, Nal
  Kalchbrenner, Ilya Sutskever, Timothy Lillicrap, Madeleine Leach, Koray
  Kavukcuoglu, Thore Graepel, and Demis Hassabis.
\newblock Mastering the game of {G}o with deep neural networks and tree search.
\newblock \emph{Nature}, 529\penalty0 (7587):\penalty0 484--489, 2016.
\newblock \doi{10.1038/nature16961}.

\bibitem[Stulp and Sigaud(2013)]{Stulp:2013}
Freek Stulp and Olivier Sigaud.
\newblock Robot skill learning: From reinforcement learning to evolution
  strategies.
\newblock \emph{Paladyn, Journal of Behavioral Robotics}, 4:\penalty0 49--61,
  2013.
\newblock \doi{10.2478/pjbr-2013-0003}.

\bibitem[Sugita and Butz(2011)]{Sugita:2011}
Yuuya Sugita and Martin~V. Butz.
\newblock Compositionality and embodiment in harmony.
\newblock In Pierre-Yves Oudeyer, editor, \emph{AMD Newsletter}, volume~8,
  pages 8--9. IEEE CIS, 2011.

\bibitem[Sugita and Tani(2005)]{Sugita:2005}
Yuuya Sugita and Jun Tani.
\newblock Learning semantic combinatoriality from the interaction between
  linguistic and behavioral processes.
\newblock \emph{Adaptive Behavior}, 13:\penalty0 33--52, 2005.

\bibitem[Sugita et~al.(2011)Sugita, Tani, and Butz]{Sugita:2011a}
Yuuya Sugita, Jun Tani, and Martin~V Butz.
\newblock Simultaneously emerging braitenberg codes and compositionality.
\newblock \emph{Adaptive Behavior}, 19:\penalty0 295--316, 2011.
\newblock \doi{10.1177/1059712311416871}.

\bibitem[Sutton and Barto(1998)]{Sutton:1998}
Richard~S. Sutton and Andrew~G. Barto.
\newblock \emph{Reinforcement learning: {A}n introduction}.
\newblock MIT Press, Cambridge, MA, 1998.

\bibitem[Tani(1996)]{Tani:1996a}
Jun Tani.
\newblock Model-based learning for mobile robot navigation from the dynamical
  systems perspective.
\newblock \emph{IEEE Transactions. System, Man and Cybernetics (Part B),
  Special Issue on Learning Autonomous Systems}, 26\penalty0 (3):\penalty0
  421--436, 1996.

\bibitem[Tani(2003)]{Tani:2003}
Jun Tani.
\newblock Learning to generate articulated behavior through the bottom-up and
  the top-down interaction processes.
\newblock \emph{Neural Networks}, 16:\penalty0 11 -- 23, 2003.

\bibitem[Tani(2017)]{Tani:2017}
Jun Tani.
\newblock \emph{Exploring Robotic Minds}.
\newblock Oxford University Press, Oxford, UK, 2017.

\bibitem[Wolpert and Flanagan(2016)]{Wolpert:2016}
Daniel~M. Wolpert and J.~Randall Flanagan.
\newblock Computations underlying sensorimotor learning.
\newblock \emph{Current Opinion in Neurobiology}, 37:\penalty0 7 -- 11, 2016.
\newblock ISSN 0959-4388.
\newblock \doi{10.1016/j.conb.2015.12.003}.

\bibitem[Wolpert and Kawato(1998)]{Wolpert:1998}
Daniel~M. Wolpert and M.~Kawato.
\newblock Multiple paired forward and inverse models for motor control.
\newblock \emph{Neural Networks}, 11:\penalty0 1317--1329, 1998.
\newblock ISSN 0893-6080.
\newblock \doi{10.1016/S0893-6080(98)00066-5}.

\bibitem[Zacks and Tversky(2001)]{Zacks:2001}
Jeffrey~M. Zacks and Barbara Tversky.
\newblock Event structure in perception and conception.
\newblock \emph{Psychological Bulletin}, 127\penalty0 (1):\penalty0 3--21,
  2001.
\newblock ISSN 1939-1455(Electronic);0033-2909(Print).
\newblock \doi{10.1037/0033-2909.127.1.3}.

\bibitem[Zacks et~al.(2007)Zacks, Speer, Swallow, Braver, and
  Reynolds]{Zacks:2007}
Jeffrey~M. Zacks, Nicole~K. Speer, Khena~M. Swallow, Todd~S. Braver, and
  Jeremy~R. Reynolds.
\newblock Event perception: A mind-brain perspective.
\newblock \emph{Psychological Bulletin}, 133\penalty0 (2):\penalty0 273--293,
  2007.
\newblock \doi{10.1037/0033-2909.133.2.273}.

\end{thebibliography}
